\documentclass{article}
\PassOptionsToPackage{table,dvipsnames}{xcolor}
\usepackage[T1]{fontenc}
\usepackage{iclr2027_conference,times}

\usepackage{amsmath,amsfonts,bm}

\def\eqref#1{equation~\ref{#1}}
\def\1{\bm{1}}

\DeclareMathAlphabet{\mathsfit}{\encodingdefault}{\sfdefault}{m}{sl}
\SetMathAlphabet{\mathsfit}{bold}{\encodingdefault}{\sfdefault}{bx}{n}

\usepackage{url}
\usepackage{amsmath}
\usepackage{amssymb}
\usepackage{enumitem}
\usepackage{booktabs}
\usepackage{wrapfig}
\usepackage{multirow}
\usepackage{makecell}
\usepackage{graphicx}
\usepackage{xcolor}
\definecolor{metrictint}{HTML}{2196F3}
\newcommand{\mc}[2]{\cellcolor{metrictint!#1}#2}
\definecolor{costtint}{HTML}{F44336}
\newcommand{\costc}[2]{\cellcolor{costtint!#1}#2}
\usepackage{xspace}
\usepackage{algorithm}
\usepackage{algpseudocode}
\usepackage{float}
\definecolor{cvprblue}{rgb}{0.21,0.49,0.74}
\usepackage[pagebackref=true,breaklinks=true,letterpaper=true,colorlinks,citecolor=cvprblue,bookmarks=false]{hyperref} % hyperlinks
\usepackage[capitalize,noabbrev]{cleveref}
\crefname{section}{Section}{Sections}
\Crefname{section}{Section}{Sections}
\crefname{table}{Table}{Tables}
\crefname{figure}{Figure}{Figures}
\crefname{algorithm}{Algorithm}{Algorithms}
\crefname{equation}{Eq.}{Eqs.}

\newcommand{\navgpt}{\textsc{NavGPT-3}\xspace}
\newcommand{\tabstyle}{\small\setlength{\tabcolsep}{4pt}\renewcommand{\arraystretch}{1.08}}
\newcommand{\contextmap}{\operatorname{Context}}

\title{\navgpt: Harnessing Context\\in a Hierarchical Navigation Runtime}

\author{\textbf{Gengze Zhou}$^{1,2\dagger}$ \quad
\textbf{Yicong Hong}$^{4}$ \quad
\textbf{Jiazhao Zhang}$^{5}$ \quad
\textbf{Xunyi Zhao}$^{1}$ \quad
\textbf{Jian Zhou}$^{1}$ \\
\textbf{Zixing Lei}$^{6}$ \quad
\textbf{Zun Wang}$^{7}$ \quad
\textbf{Chongyang Zhao}$^{8}$ \quad
\textbf{Xionghui Chen}$^{5}$ \\
\textbf{Stephen Gould}$^{2,3}$ \quad
\textbf{Anton van den Hengel}$^{1,2}$ \quad
\textbf{Qi Wu}$^{1,2\dagger}$ \\
$^1$AIML, Adelaide University; $^2$Metacognition; $^3$ANU; \\
$^4$Roblox; $^5$PKU; $^6$SJTU; $^7$UNC Chapel Hill; $^8$UNSW \\
$^\dagger$Correspondence to: \texttt{\{gengze.zhou, qi.wu01\}@adelaide.edu.au} \\
\makebox[\textwidth][l]{\normalfont\small Project Page: \urlstyle{same}\url{https://metacognitionai.github.io/NavGPT3/}} \\
}

\iclrfinalcopy
\begin{document}
\raggedbottom

\maketitle
\fancyhead{}
\fancypagestyle{navgpttitle}{%
  \lhead{\raisebox{-2pt}[0pt][0pt]{\includegraphics[height=14pt]{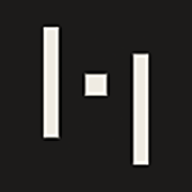}}\hspace{0.5em}{\sffamily\bfseries Metacognition}\hspace{1.2em}\raisebox{-2pt}[0pt][0pt]{\includegraphics[height=14pt]{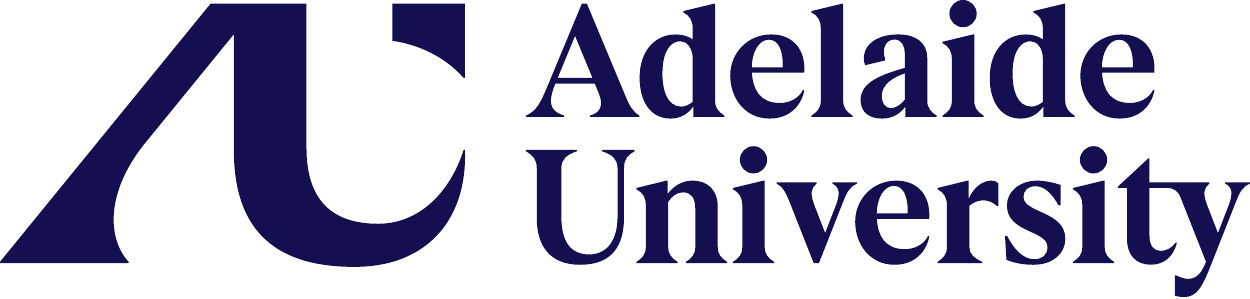}}}%
}
\thispagestyle{navgpttitle}

\begin{abstract}
Language models trained with long-horizon agentic reinforcement learning can generalize knowledge through reasoning, express precise actions, and pursue goals over many steps, raising the ceiling on what an embodied agent can understand and decide. Physical interaction, however, remains the domain of action policies, which provide dense, low-latency control. We present \navgpt, a harness that connects the two models, with an OS-like runtime built above it: reasoning, acting, and monitoring run as threads with their own context, tools, and permissions, while the runtime schedules them and decides which thread controls the robot's motion, so that the robot can react to sudden real-world events through interruption and thread switching. Beneath it, our action policy NavGPT VLA, trained on 19.28M examples, allocates visual tokens using codec allocation, in proportion to scene change; its 8B model alone reaches 76.94 SR on R2R-CE and leads RxR-CE with 82.39 SR. With the complete harness, \navgpt sets the state of the art on R2R-CE (81.51 SR) and, for the first time, brings an autonomous agent to human level: on RxR-CE it matches human followers in success (90.43 vs.\ 90.4 SR) and path fidelity (78.47 vs.\ 77.7 nDTW) at 1\,min\,22\,s per episode, versus roughly 3\,min for a human. We comprehensively ablate the harness design and the interaction between the two models, showing how tools and the action policy shape the path from language-model reasoning to physical control: when NavGPT VLA executes the route, the reasoning loop shortens and the system's minimum reaction time falls from 3--19\,s per language-model decision to 0.5--1\,s per action-policy step (1--2\,Hz). These results show that designing this embodied interface is central to connecting frontier language-model intelligence with low-level physical control. We will release all models, code, and evaluation records.
\end{abstract}

\begingroup
\setlength{\intextsep}{2pt}
\begin{figure}[H]
\centering
\includegraphics[width=\linewidth]{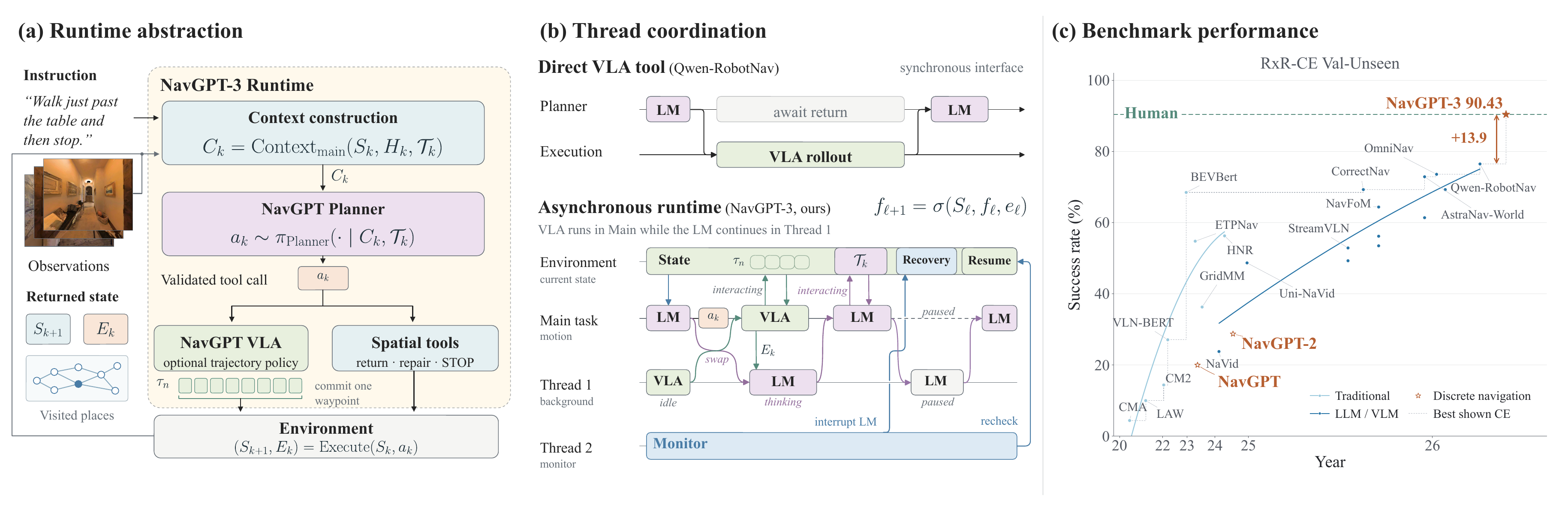}
\caption{Overview of \navgpt: (a) runtime abstraction with the planner, VLA, and spatial tools; (b) synchronous tool use versus asynchronous thread coordination; and (c) benchmark performance. \textbf{\navgpt reaches human performance on RxR-CE}.}
\label{fig:teaser}
\end{figure}
\endgroup

\section{Introduction}
\label{sec:intro}

Language models trained with long-horizon agentic reinforcement learning generalize knowledge through reasoning, express actions as tool calls and code, and pursue goals over many steps~\citep{gpt6astra,claudeopus5,claudefable51,qwen38max}. In Vision-and-Language Navigation (VLN), successive generations of language models absorb more of what navigation systems once built as separate modules: commonsense grounding, spatial description, progress estimation, and long-horizon tool use. Execution is the exception. It runs on the world's clock rather than the model's, and control must proceed at a rate that step-by-step reasoning cannot sustain. The world also does not pause while a model thinks: observations arrive during inference, motion continues past the decision that began it, and a hazard may demand a stop before any reasoning finishes. An action policy is also far smaller and weaker at following instructions than the model that plans, so it drifts over a long rollout and must report back evidence the reasoner can check. Existing systems address part of this by combining reasoning and execution hierarchically, with a planner issuing subgoals to a learned controller~\citep{gao2025vlaos,hu2026hivla,chiang2024mobilityvla,route2step2026}, and recent agents add graph memory and backtracking~\citep{hamvln2026} or verification and recovery skills~\citep{embodiedskills2026}.

Connecting these abilities to physical control therefore requires a complete interaction loop: grounded context for deciding, spatial tools for acting, persistent state for remembering, and execution evidence for revising a route. A \emph{harness} shapes model behavior through this loop, defining the context models receive, the tools they can call, and the state and feedback retained across calls~\citep{zhong2026harness,deepseekharness2026}. Real robots add a coordination requirement: reasoning, execution, and monitoring must be able to progress asynchronously. A \emph{runtime} above the harness manages these activities as threads and controls which thread may move the robot.

We present \navgpt, which combines a complete navigation harness with an OS-like runtime. The harness brings context management, spatial tools, persistent route state, and model execution into one embodied interface. The Planner can act through those tools or delegate execution of a navigation instruction to NavGPT VLA. Returned route images, execution feedback, an observed map, and numbered visited places make the resulting rollout inspectable and revisable: the Planner can return to a visited place, repair the route locally, delegate again, and explicitly decide when the task is complete. We developed these harness interfaces through recursive self-improvement (RSI), using automated research iterations over tool definitions and visual prompts (\Cref{app:tool_development}). The runtime builds on these interfaces to schedule threads and control \emph{motion authority}: review can overlap execution, and monitoring can interrupt motion without waiting for reasoning (\Cref{fig:runtime_sequence}). Tool-capability and visual-context studies examine the harness design alongside Planner--VLA interaction (\Cref{sec:planner_models,sec:runtime_ablation}).
Beneath the harness, NavGPT VLA predicts eight waypoints from the instruction and visual history. Because the visual context is bounded, we use codec allocation from video coding: the first observation of each view is a full reference (I-frame), and later observations (P-frames), receive visual tokens in proportion to how much they change~\citep{onevisionencoder,onevision2}.

Across benchmarks, the 8B NavGPT VLA alone leads published methods on RxR-CE (82.39 SR); under the runtime, \navgpt sets the state of the art on R2R-CE (81.51 SR) and, on RxR-CE, matches human followers in both success (90.43 vs.\ 90.4 SR) and path fidelity (78.47 vs.\ 77.7 nDTW). The same runtime transfers to object-goal navigation and embodied question answering (\Cref{tab:sota,tab:objectnav,tab:eqa}). Ablating the harness tools and the VLA interaction shows how this interface governs the path from reasoning to control: given good tools, a strong language model navigates well by itself but decides only once every 3--19\,s, whereas letting the VLA execute the route lowers the system's minimum reaction time to 0.5--1\,s per step (1--2\,Hz; \Cref{tab:planner_models}).

\section{Related Work}
\label{sec:related}

\noindent\textbf{Agentic Vision-and-Language Navigation with LLMs.} VLN requires an agent to follow natural-language instructions in unseen environments, first on discrete navigation graphs \citep{anderson2018vision,anderson2020rxr} and then in continuous environments \citep{krantz2020beyond}. NavGPT initiated the line of work asking how much of navigation a LLM can carry on its own~\citep{zhou2024navgpt}: given observations in text, the language model reasons about instruction progress and decides each move zero-shot, without any training on action data. Later navigators extend this recipe with maps, discussion, and open-vocabulary grounding~\citep{chen2024mapgpt,pan2023langnav,long2023discuss,qiao2025opennav}, and recent agents let the model act on the embodiment directly through code~\citep{zhou2026takecontrol}. A parallel line of work targets physical execution, tuning Vision-Language Models (VLMs) into VLA policies that provide dense, low-latency control while retaining the generalization of the VLM~\citep{zhang2024navid,zhang2024uni,cheng2024navila,wei2025streamvln,wei2025dualvln,cai2025internvla,zhang2025embodied,abotN0,qwennav,robostralnavigate}. NavGPT-2 pursues both at once, emphasizing the role of reasoning in language models while training a policy that acts faster and more accurately, with the language and action heads decoupled over one shared representation~\citep{zhou2024navgpt2}. However, such training cannot perfectly align the reasoning and the actions. NavGPT-3 combines the two at the system level, connecting a reasoning model and a trained VLA policy through a harness with an OS-like runtime built above it.

\noindent\textbf{Language models over learned policies.} Hierarchical systems pair a planning model with a learned controller in navigation \citep{chiang2024mobilityvla,route2step2026,hamvln2026} and manipulation \citep{gao2025vlaos,hu2026hivla,guo2026planar,li2026roboclaw,yang2026hivla,embodiedskills2026}, and harness engineering treats context, tools, and state as the unit of design \citep{zhong2026harness,deepseekharness2026}. These systems already cover graph memory, backtracking, verification, and recovery. The closest system, NavMCP, retains trajectory evidence across calls to a navigation foundation model in order to answer questions \citep{lei2026navmcp}. NavGPT-3 instead uses returned evidence to revise an instruction-following route: after a delegated VLA rollout, the Planner can inspect the route, return to any visited place, repair locally, and decide completion.

\noindent\textbf{Visual context in navigation policies.} Navigation policies must fit long, multi-view observation histories into a bounded visual budget. Existing policies compress history by merging or pruning visual tokens \citep{zhang2024uni,wei2025streamvln} or distribute a fixed budget across time and views \citep{qwennav}. NavGPT VLA allocates tokens in proportion to scene change, as codec-aligned video encoders do \citep{onevisionencoder,onevision2}.

\section{Method: Harness, Runtime, and VLA Co-Design}
\label{sec:method}

\navgpt co-designs a complete harness between a reasoning model and a navigation policy: context and state management, spatial tools, validated execution, structured returns, route repair, and an explicit decision to end the task. A runtime built above the harness controls its threads. Part I describes these two layers; Part II describes the VLA beneath them.

\subsection{Part I: The NavGPT-3 Harness and Runtime}
\label{sec:harness_method}

\subsubsection{Harness and Runtime Abstraction}
\label{sec:formulation}

\begin{figure}[t]
\centering
\includegraphics[width=\linewidth]{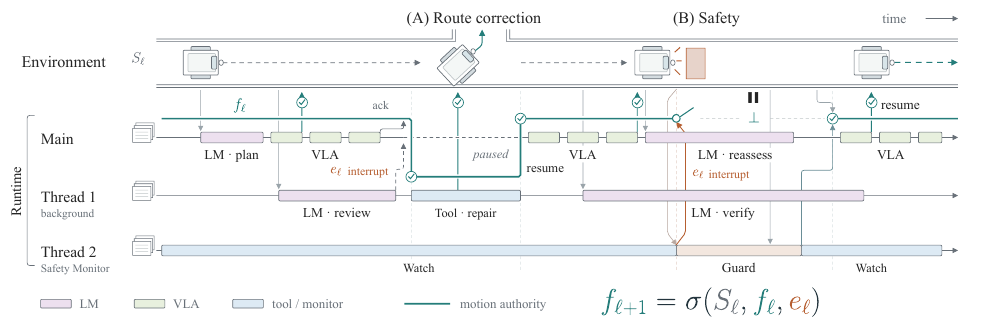}
\caption{Illustrative foreground-control patterns in the \navgpt Runtime. Thin bars mark concurrent activity; the teal path denotes motion authority. (A) A route-review interrupt pauses main execution for correction, followed by resumption. (B) A safety interrupt revokes authority while inference continues; resumption requires fresh state and runtime authorization.}
\label{fig:runtime_sequence}
\end{figure}

The harness shapes model behavior through the context, state, and tool interfaces that connect the Planner, the VLA, and the environment (simulator or robot). It builds the observations and history the models see, keeps spatial references and annotations, checks tool requests, executes permitted calls, and returns evidence for the next decision. The Planner decides which step to take next and when the task is complete; the VLA optionally provides high-frequency execution. Built above the harness, the OS-like runtime controls logical threads, their lifecycle and permissions, and motion authority. \Cref{fig:runtime_sequence} illustrates this coordination through route review and a protective monitor.

\subsubsection{Context, Spatial Tools, and Execution Returns}
\label{sec:context_compiler}
\label{sec:tools}

The harness turns observations and interaction history into decision context, and exposes spatial tools that make this context actionable. Because invoking the Planner at every sensorimotor update is impractical, the harness additionally exposes a VLA for extended navigation through a higher-frequency perception--action loop. Invoking the VLA is optional: the Planner decides whether to use it or make all navigation decisions itself through the available tools. At invocation $k$, it selects
\begin{equation}
    a_k\sim\pi_{\mathrm{Planner}}(\cdot\mid C_k,\mathcal{T}_k),
    \label{eq:planner_choice}
\end{equation}
where $C_k$ is the harness-provided context, $\mathcal{T}_k$ the permitted tool set, and $a_k$ a call including its arguments. The harness checks the selected call and executes it if the runtime permits; within these limits, the Planner determines the navigation strategy.

The interface groups operations into observation, instruction delegation, spatial revision, and semantic completion. Observation returns current views and observed geometry; delegation returns route keyframes and execution status; revision resolves visited-place references or local corrections into motion; annotation and termination preserve meaning and make completion explicit. These tools and their visual returns were iterated through RSI (\Cref{app:tool_development}). Their shared interface makes extended execution inspectable and revisable: a VLA stop ends one operation, while the Planner retains final task-stop authority (\Cref{app:runtime_interfaces}).

\subsubsection{Execution Coordination and Motion Authority}
\label{sec:collaboration}

Threads perform inference and receive observations at their own cadences, subject to their permissions, so activity on different rows of \Cref{fig:runtime_sequence} overlaps in time. The runtime controls which thread holds \emph{motion authority}, the permission to issue task motion (teal path). Let $f_\ell\in\mathcal{J}_\ell\cup\{\bot\}$ denote the foreground thread at event index $\ell$, where $\mathcal{J}_\ell$ is the thread set and $\bot$ denotes no holder. Control transfer follows
\begin{equation}
    f_{\ell+1}=\sigma(S_\ell,f_\ell,e_\ell),
    \label{eq:foreground_transfer}
\end{equation}
where $S_\ell$ is shared system state, including the environment (top row), $e_\ell$ an incoming event such as an interrupt (orange), and $\sigma$ the runtime's control-transfer rule, which uses preassigned event priorities to retain, transfer, or revoke authority. Protective events have highest priority; episode limits take precedence over routine review, and stale events cannot restore an obsolete operation (\Cref{app:priority_policy}). A handoff preserves thread contexts and takes effect only after runtime authorization and execution acknowledgement (checkmarks). Event order $\ell$ does not impose a common inference clock.

For a task-motion operation $a$ proposed by thread $j$, foreground eligibility is
\begin{equation}
    \operatorname{eligible}_{\ell}(j,a)=\mathbf{1}\!\left[j=f_\ell\ \land\ a\in\mathcal{P}_j\right],
    \label{eq:foreground_eligibility}
\end{equation}
where $\mathcal{P}_j$ is the thread's permitted operation set. Eligibility is necessary, not sufficient: execution also checks the current state and permissions (\Cref{app:interrupt_protocol}). Motion authority does not grant exclusive computation or sensing, and calling the Planner or VLA within a thread requires no switch.
This mechanism supports both concurrent progress and synchronization. For example, in route correction (\Cref{fig:runtime_sequence}A), the VLA holds authority while the background thread reviews its progress; the review raises $e_\ell$, and after the VLA acknowledges, authority passes to a tool repair while the main thread pauses, then returns so the VLA resumes. In the safety case (B), the monitor's guard revokes authority, $f_{\ell+1}=\bot$, without awaiting LLM inference or the routine acknowledgement, while reasoning continues; the main thread resumes only with fresh state and authorization.

\begin{figure}[!t]
\centering
\includegraphics[width=\linewidth]{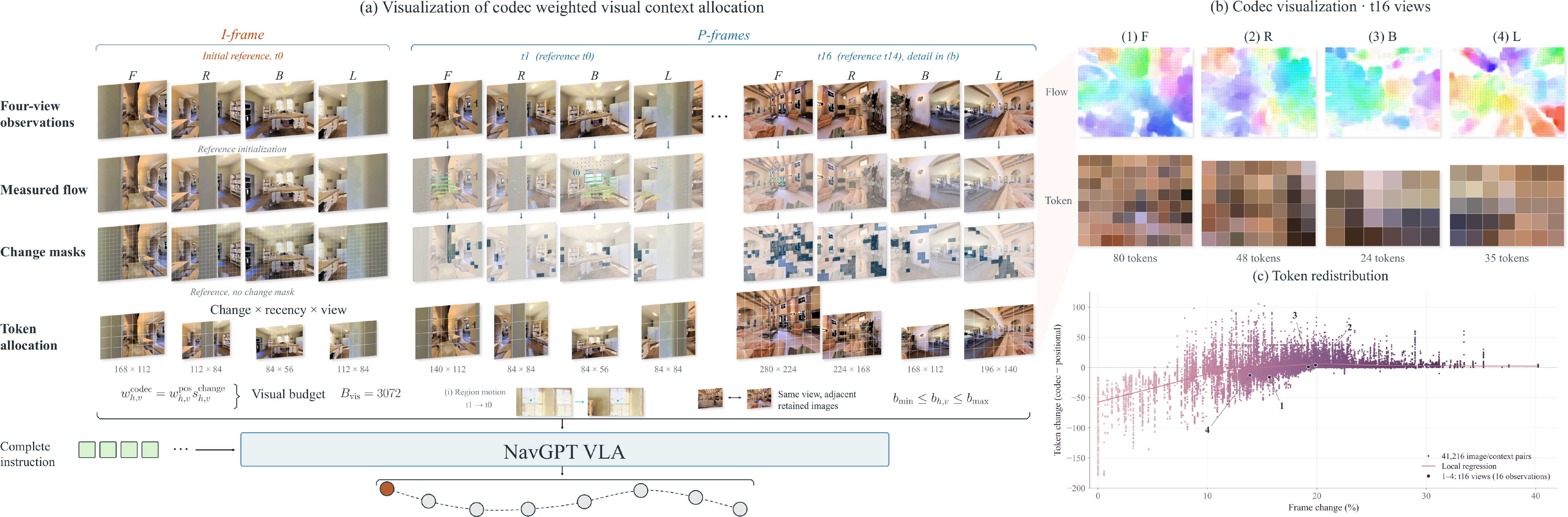}
\caption{\textbf{Codec-weighted visual-context allocation.} (a) Change, recency, and view jointly determine whole-image resolution for a four-view history, with an initial I-frame and subsequent P-frames. (b) Optical-flow diagnostics and resulting token grids for the four views at $t_{16}$. (c) Allocation change relative to positional weighting versus RGB change $100\delta_{h,v}$ for 41,216 replayed image/context pairs from 13 recorded histories at $B_{\mathrm{vis}}=3072$. Curve: local regression; numbers match (b).}
\label{fig:codec_measured}
\end{figure}

\subsection{Part II: NavGPT VLA}
\label{sec:vla_training}

We train NavGPT VLA as a \emph{versatile navigation foundation model}: one vision-language-action backbone that learns from several navigation tasks, camera configurations, and embodiments. The 4B and 8B models are trained on instruction following, point-goal and object-goal navigation, target tracking, and outdoor driving, together with embodied question answering and visual grounding. Within NavGPT-3, this policy executes the navigation instructions issued by the Planner.

\subsubsection{Trajectory Policy and Visual-Context Allocation}
\label{sec:vla_policy}
\label{sec:codec}

NavGPT VLA starts from Qwen3-VL-Instruct and adds an MLP action head. At VLA step $n$, it uses the provided instruction and up to 16 retained observations, each with front, right, back, and left views, to predict an ordered trajectory $\tau_n=((x_{n,r},y_{n,r},\theta_{n,r}))_{r=1}^{8}$. Waypoint index $r$ runs over eight cumulative targets in the current agent frame: $x$ points forward and $y$ points left (both in metres), and $\theta$ is relative heading in radians. Both 4B and 8B variants share this output format; execution commits only the first waypoint before the next prediction. Training combines waypoint regression with next-token supervision as detailed in \Cref{app:vla_training}.

The VLA sees up to $N_{\mathrm{hist}}$ retained observations of $N_{\mathrm{view}}$ views (F, R, B, L in \Cref{fig:codec_measured}a) under a visual-token budget $B_{\mathrm{vis}}$; image $(h,v)$, view $v$ of observation $h$, receives $b_{h,v}$ tokens. Our reimplementation of Qwen-RobotNav's task-adaptive allocation~\citep{qwennav} uses position alone: a weight $w^{\mathrm{pos}}_{h,v}$ reflects recency and view direction, and an allocator converts the weights into integer counts with $b_{\min}\le b_{h,v}\le b_{\max}$ (\Cref{app:codec_protocol}). Position records when and where an image was taken, but not whether it shows anything new.

Codec allocation adds a change score, inspired by I/P-frame video coding (\Cref{fig:codec_measured}a). Within each view, we compare $64\!\times\!64$ RGB thumbnails of consecutive retained images: the first image ($t_0$) is an I-frame with full score, and later P-frames are scored by their mean absolute pixel difference $\delta_{h,v}$,
\begin{equation}
    \delta_{h,v}=\frac{\lVert\operatorname{thumb}(I_{h,v})-\operatorname{thumb}(I_{h-1,v})\rVert_1}{255\cdot3\cdot64^2},\qquad
    s^{\mathrm{change}}_{h,v}=\begin{cases}1,&\text{I-frame},\\ \operatorname{clip}(5\delta_{h,v},0.05,1),&\text{P-frame},\end{cases}
    \label{eq:codec_score}
\end{equation}
The codec weight $w^{\mathrm{codec}}_{h,v}=w^{\mathrm{pos}}_{h,v}s^{\mathrm{change}}_{h,v}$ thus combines change, recency, and view (bottom row of \Cref{fig:codec_measured}a); the allocator and token total are unchanged, and a constant score recovers the positional rule. Whether the score can act depends on the budget-to-capacity ratio $\rho_{\mathrm{vis}}=B_{\mathrm{vis}}/(N_{\mathrm{hist}}N_{\mathrm{view}}b_{\max})$: at $\rho_{\mathrm{vis}}\ge1$ every image can reach $b_{\max}$ and the two rules coincide, while below one tokens shift toward changing views. Training randomizes $B_{\mathrm{vis}}\in[1024,4096]$ and the allocation limits so that non-trivial histories enter this active regime (\Cref{tab:saturation}).

At $t_{16}$ (\Cref{fig:codec_measured}b), F, R, B, and L receive 80, 48, 24, and 35 tokens; the flow fields and change masks only visualize inter-frame differences, as allocation uses the thumbnail score in \Cref{eq:codec_score}. Across recorded histories (\Cref{fig:codec_measured}c), codec allocation takes tokens from low-change images and moves them to images that change.

\subsubsection{Training Data Taxonomy and Mixture}
\label{app:data_mixture}

\begin{figure}[htbp]
  \centering
  \includegraphics[width=\linewidth]{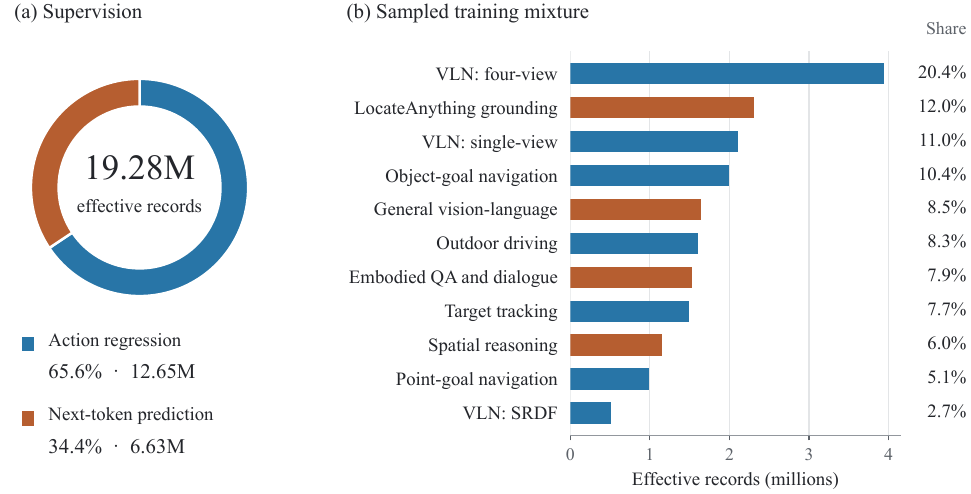}
  \caption{\textbf{NavGPT VLA training mixture.} Supervision split and effective records by data family.}
  \label{fig:data_mixture}
\end{figure}

A foundation action model inherits broad visual and linguistic knowledge from its pretrained backbone, but how reliably it acts depends on the range of situations its action data cover, and adding demonstrations of a single task does little to cover the edge cases a deployed policy meets. We therefore build the NavGPT VLA corpus around a taxonomy of coverage rather than the volume of any one benchmark. The taxonomy has four axes: the capability a task demands, the embodiment and observation interface through which it is performed, the conditions under which it occurs, and the visual-language understanding it presupposes. The first three are supervised through actions and the fourth through language. Every navigation source, whatever its origin, is converted to the same eight-waypoint target in the agent frame (\Cref{app:vla_training}), so coverage gained on any axis trains one shared action space rather than separate task-specific policies (\Cref{tab:data_mixture,fig:data_mixture}).

\paragraph{Capability.}
We span four capabilities that place progressively broader demands on the policy. Instruction following grounds guidance in motion, from step-by-step language to compact point goals that give only a direction or a relative coordinate. Active exploration, realized by object-goal navigation, removes the route altogether: the agent must search a partially observed scene and infer where a target is likely to be. Dynamic interaction, realized by target tracking, adds agents whose motion must be anticipated as it unfolds, while the target's identity is kept among distractors. Cross-embodiment navigation, realized mainly by outdoor driving data from OpenScene~\citep{openscene2023} and nuScenes~\citep{caesar2020nuscenes}, carries the same spatial planning to a vehicle in traffic, with a longer perception range, a different motion scale, and stricter safety constraints. Along this sequence, uncertainty about the environment, dependence on history, and embodiment diversity roughly increase. The capabilities are not disjoint: harder regimes reuse simpler skills while adding their own perception and control demands. Joint training over this overlap is intended to yield spatial planning that transfers across task families instead of being fit to each one separately.

\paragraph{Embodiment and observation interface.}
A policy that serves several platforms must act from whatever cameras a platform carries. The most economical way to cover this axis is to re-render existing episodes rather than collect new ones. Each R2R-CE and RxR-CE training episode~\citep{krantz2020beyond,anderson2020rxr} therefore appears both as a four-view panorama from front, right, back, and left cameras and as a front-camera sequence, with identical waypoint targets, so the policy learns one behavior from surround and forward-only input. Front-view trajectories from VLNVerse~\citep{lin2025vlnverse}, robot-navigation demonstrations from other platforms such as RoboCasa~\citep{nasiriany2024robocasa}, and the driving data~\citep{openscene2023,caesar2020nuscenes} extend the same interface to further simulators, bodies, and camera rigs. Within simulation, object-goal episodes start from random poses with randomized camera height and field of view, so a single source already spans several sensor placements.

\paragraph{Conditions and edge cases.}
Within a task, failures concentrate where the training distribution is thin: unfamiliar phrasing, appearance, and scenes; states off the expert path; and rare actions. We widen each of these while keeping the supervision fixed where possible. For language and appearance, we refine the instructions in R2R-CE and RxR-CE episodes with Gemini 3.1 Pro~\citep{gemini31pro} and with subsets in VLN and tracking data whose rendered observations from habitat are refined with Qwen-Image-Edit~\citep{wu2025qwenimage} to look closer to real camera images. These variants change what the policy reads or sees but not the route it must take. For scenes, SRDF~\citep{wang2025srdf} supplies instruction--trajectory pairs in the HM3D environments~\citep{wang2023scaling}, which span many more scenes than the R2R training split in MP3D. We rebuild these data in two ways. First, the original instructions are often inaccurate, so we regenerate all of them with Gemini 3.1 Pro~\citep{gemini31pro}, which rewrites each instruction from the trajectory video, the shortest-path video to the same goal, and the original instruction. Second, for state coverage we keep the DAgger rollouts alongside the shortest-path routes but change their supervision. The original data label each visited state with the rollout's own next step, a pseudo action that need not move the agent toward the goal; we instead label every visited state with the direction of the shortest path from that position to the goal. The relabeled rollouts thus cover off-path states that shortest-path demonstrations never visit and supervise the recovery from each of them. For actions, geometry-aware resampling reduces the forward-motion bias of raw trajectories and raises the share of turns and stops (\Cref{app:vla_training}). The point-goal data additionally vary how fast the agent moves and how open the space around it is. Standard episodes advance about 0.25\,m per step, whereas low-speed episodes advance about 0.1\,m, so the same displacement unfolds over more than twice as many observations and the policy must judge progress from much smaller changes between frames. Open-space episodes place goals at round metric offsets, such as 4\,m ahead and 1\,m to the left, in open areas of a scene, where walls and corridors no longer channel the route and the policy must reach the goal from the coordinates alone.

\paragraph{Visual-language understanding.}
Action supervision teaches where to move, but not what is in a scene or where a named object appears in the image. Language-supervised data, trained with next-token prediction only, cover this axis, and among them we found grounding data to be the most important. We draw it from LocateAnything-Data~\citep{wang2026locateanything}, a corpus of 138M language queries with 785M boxes over 12M images. Our 19 subsets contribute 2.31M effective records, 12.0\% of the complete mixture and second only to four-view instruction following. They span object detection (40.0\% of these records, including egocentric, part-level, person, and driving scenes), referring-expression grounding (18.2\%), point and affordance localization (13.7\%), GUI element grounding (12.4\%), text localization (11.0\%), and document layout (4.6\%). Every example pairs a language query with the boxes or points it refers to, which trains the backbone to map words to precise image locations across natural, egocentric, robotic, driving, and screen images. Navigation needs the same ability whenever an instruction names a landmark or a target object, and waypoint supervision alone does not provide it, since it ties language to motion rather than to explicit image positions. The remaining language-supervised data add spatial reasoning from LLaVA-OneVision-2~\citep{onevision2}, question answering and dialogue on navigation episodes, and general vision-language instruction data that retain the backbone's broader abilities while it learns to act.

\paragraph{Mixture composition.}
The complete mixture draws on 83 sources grouped into 11 families (\Cref{tab:data_mixture}) and contains approximately 19.28M effective records per epoch. As shown in \Cref{fig:data_mixture}, 65.6\% of these records supervise action regression and 34.4\% supervise next-token prediction only. Spatial reasoning and grounding together contribute approximately 18\% of the records, and general vision-language data 8.5\%.

\paragraph{Sampling and counting.}
Most sources are used in full, and several large sources are subsampled (\Cref{tab:data_mixture}): the SRDF navigation sources keep 15\%, the SRDF-based dialogue sources 10\%, the two largest general vision-language sources 30\%, and the largest spatial-reasoning and LocateAnything sources between 5\% and 60\%. For source $d$ with $n_d$ records, sampling fraction $s_d\in(0,1]$, and repetition count $r_d$, the effective size $\widetilde{N}_d=n_ds_dr_d$ is the number of samples it contributes to one epoch, and a family's share is its summed effective size divided by $\sum_d\widetilde{N}_d$. Effective records count training samples, not unique trajectories, scenes, or datasets: the four-view and single-view R2R/RxR data render the same episodes, the refined-instruction and refined-observation variants reuse their routes, and SRDF builds on HM3D environments.
\begin{table}[htbp]
  \centering
  \small
  \caption{\textbf{Training-data mixture.} Sizes are effective records per epoch, in millions.}
  \label{tab:data_mixture}
  \setlength{\tabcolsep}{4pt}
  \renewcommand{\arraystretch}{1.08}
  \begin{tabular*}{\linewidth}{@{\extracolsep{\fill}}lrcr@{}}
    \toprule
    Family & Sources & Sampling & Size (M) \\
    \midrule
    \multicolumn{4}{@{}l}{\textit{Action regression}} \\
    VLN: four-view & 7 & 100\% & 3.938 \\
    VLN: single-view & 5 & 100\% & 2.113 \\
    VLN: SRDF & 2 & 15\% & 0.516 \\
    Point-goal navigation & 5 & 100\% & 0.984 \\
    Object-goal navigation & 1 & 100\% & 2.000 \\
    Target tracking & 6 & 100\% & 1.486 \\
    Outdoor driving & 5 & 100\% & 1.608 \\
    \midrule
    \multicolumn{4}{@{}l}{\textit{Next-token prediction}} \\
    Embodied QA and dialogue & 12 & 10--100\% & 1.532 \\
    General vision-language & 13 & 30--100\% & 1.640 \\
    Spatial reasoning & 8 & 25--100\% & 1.149 \\
    LocateAnything grounding & 19 & 5--100\% & 2.311 \\
    \midrule
    Total & 83 & & $\approx$19.28 \\
    \bottomrule
  \end{tabular*}
\end{table}

\subsubsection{Data Processing and Training}
\label{sec:data_processing}
\label{app:vla_training}

Navigation sources are converted to a common multimodal format containing an instruction, an ordered image sequence, and an eight-waypoint target. Four-view Habitat observations are rendered in front/right/back/left order, while single-view sources retain the front camera; the first and current observations are always retained and intermediate history is sampled up to 16 steps. Habitat discrete actions are simulated as 0.25\,m forward motion, 15-degree turns, or padded stops and transformed into cumulative agent-local coordinates normalized by $(1.0,0.433,2.094)$ for $(x,y,\theta)$. Geometry-aware resampling reduces the raw forward-action bias and increases turns and stops. Embodied navigation QA, general vision-language, spatial-reasoning, and grounding examples use the same image-and-conversation format but omit the trajectory target and train only the next-token objective.

Both model sizes are initialized from Qwen3-VL-Instruct and trained for one epoch on the complete mixture with the same global batch size, schedule, codec allocation, and action output; model and action-head learning rates are $5\times10^{-6}$ and $10^{-3}$, respectively (\Cref{tab:model_config}). The ablations in \Cref{app:codec_ablation,app:vla_scaling} train additional models on smaller cumulative scopes of this mixture.

\begin{table}[H]
  \centering
  \small
  \caption{\textbf{Matched VLA training configurations.} BS: global batch size; AH LR: action-head learning rate.}
  \label{tab:model_config}
  \setlength{\tabcolsep}{5pt}
  \begin{tabular}{lrrrrrr}
    \toprule
    \textbf{Backbone} & \textbf{GPUs} & \textbf{BS} & \textbf{Steps} & \textbf{Length} & \textbf{LR} & \textbf{AH LR} \\
    \midrule
    Qwen3-VL-4B & 32 & 256 & $\sim$75.3k & 8,192 & $5{\times}10^{-6}$ & $10^{-3}$ \\
    Qwen3-VL-8B & 32 & 256 & $\sim$75.3k & 8,192 & $5{\times}10^{-6}$ & $10^{-3}$ \\
    \bottomrule
  \end{tabular}
\end{table}

\paragraph{Supervision and masking.}
Training combines trajectory regression with vision-language co-training~\citep{qwennav}, using the coordinate normalization and training configuration above. The MLP action head predicts a 24-dimensional vector containing eight $(x,y,\theta)$ waypoints. For a minibatch with navigation examples $\mathcal{B}_{\mathrm{nav}}$, its loss is
\begin{equation}
\mathcal{L}_{\mathrm{nav}}=\frac{1}{24|\mathcal{B}_{\mathrm{nav}}|}\sum_{i\in\mathcal{B}_{\mathrm{nav}}}\sum_{r=1}^{8}\left\|\hat{\mathbf{a}}_{i,r}-\mathbf{a}_{i,r}\right\|_2^2,\qquad \mathcal{L}=\mathcal{L}_{\mathrm{nav}}+\mathcal{L}_{\mathrm{NTP}}.
\end{equation}
Both losses have unit weight. Position and heading have equal weight after normalization; the coordinate scales therefore determine their relative weight in physical units. The fixed eight-waypoint target includes padded stop positions near trajectory ends, and the MLP loss applies no additional terminal-waypoint mask. Navigation examples are excluded from next-token supervision. Conversely, QA and grounding examples have no action target and contribute only cross-entropy over labeled answer tokens; prompt and padding tokens are excluded from the loss. Their empty trajectory targets are masked out rather than learned as zero-motion targets. An absent example type contributes zero to its corresponding loss.

\section{Experiments}
\label{sec:experiments}

This section evaluates the \navgpt harness, the collaboration with the VLA, and the complete system. In \Cref{app:tool_development}, we first show the development process of the harness tools and of the visual context shown to the model by automated recursive self-improvement on a subset of training episodes. In \Cref{sec:planner_models}, we evaluate the resulting tools: we add the VLA, Map, and Backtrack tool groups to the basic tools for two Planners and measure accuracy, cost, and reaction time. In \Cref{sec:runtime_ablation}, we vary how tool returns are presented to the Planner, and in \Cref{sec:interruption}, how often the Planner checks VLA execution. In \Cref{sec:system_results}, we compare the complete system with published methods on R2R-CE and RxR-CE and transfer it to object-goal navigation and embodied question answering. In \Cref{app:codec_ablation,app:vla_scaling}, we study NavGPT VLA on its own: its codec allocation and its scaling with data and model size. Finally, in \Cref{app:deployment}, we deploy the system on a Unitree Go2 robot, and in \Cref{app:qualitative}, we show qualitative episodes.

For the navigation performance, we evaluate \navgpt on the full Val-Unseen splits of R2R-CE and RxR-CE~\citep{krantz2020beyond,anderson2018vision,anderson2020rxr}, on object-goal navigation~\citep{batra2020objectnav} in MP3D~\citep{chang2017matterport3d} and HM3D v2~\citep{ramakrishnan2021hm3d}, and on embodied question answering~\citep{ren2024explore,zhai2025memoryeqa,jiang2025express}; ablation studies use a fixed R2R subset of 100 episodes across ten MP3D scans~\citep{qiao2025opennav}. We report success rate (SR), success weighted by path length (SPL), normalized dynamic time warping (nDTW), oracle success (OSR), and navigation error (NE), with accuracy and normalized steps for question answering. Unless varied, harness studies use the 8B NavGPT VLA. Each Planner runs inside its provider's scaffold, Codex for GPT-6 Astra and Claude Code for Claude Opus 5, which assembles messages and manages the conversation, including automatic compression of long contexts. \navgpt replaces the system prompt, tools, observations, spatial state, and stopping rule; differences between the two frameworks are described in \Cref{app:experiment_protocols}.

\subsection{Harness Development Through Recursive Self-Improvement}
\label{app:tool_development}

\begin{figure}[t]
\centering
\includegraphics[width=\textwidth]{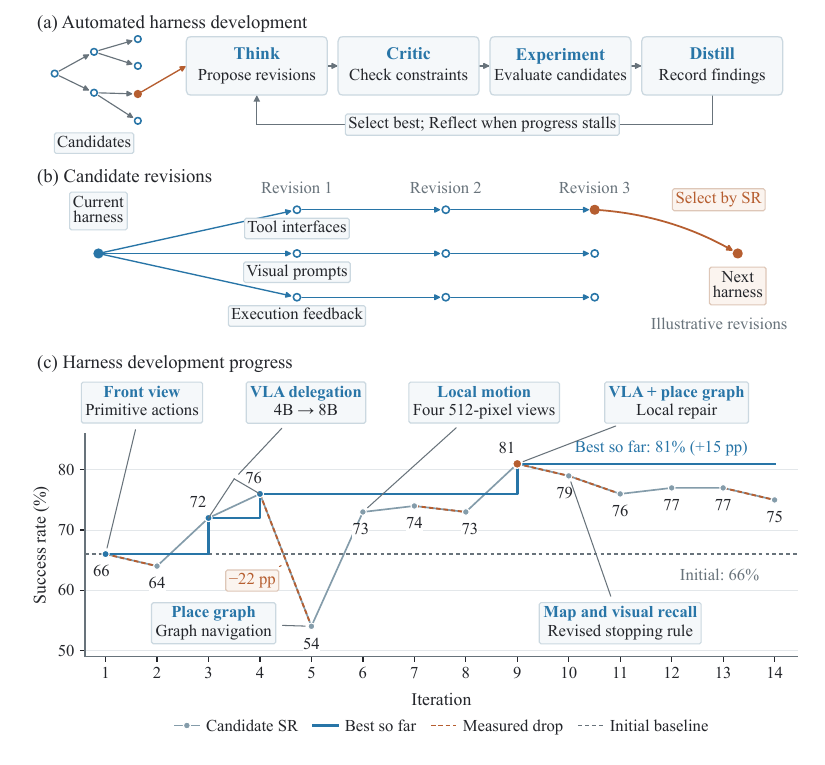}
\caption{Harness development through recursive self-improvement: the research loop, candidate revisions, and success-rate progression.}
\label{fig:rsi_loop}
\end{figure}

We formulate recursive self-improvement (RSI) of the harness as a KDLoop-style tree search over executable harness versions \citep{zhou2026agentcanvas}. Each node specifies the Planner's system prompt, the available tools, the persistent spatial state, and the formats in which observations and execution results are shown to the Planner; each edge is a candidate revision. \textsc{Think} proposes edits using the current harness and accumulated episode evidence; \textsc{Critic} checks them against interface constraints and earlier failures; \textsc{Experiment} implements and scores the surviving branches; and \textsc{Distill} records confirmed findings, failed edits, and unresolved hypotheses. The highest-scoring candidate becomes the next harness, while \textsc{Reflect} revisits whole categories of changes when progress stalls. Search operates at development time: the chosen interface is fixed during evaluation, without online Planner or VLA weight updates. \Cref{fig:rsi_loop}(a) summarizes the research loop; panel (b) illustrates three categories of changes, to tool interfaces, visual context, and execution results, which branch from a common harness, are revised, and merge the selected branch into the next best harness.

The search evaluates candidate harnesses with Claude Opus 5 on one fixed 100-episode subset sampled from the R2R-CE training split, whose scenes are disjoint from Val-Unseen, using success rate as the selection signal and episode traces to diagnose failures. \Cref{fig:rsi_loop}(c) shows the 14 harness configurations that were evaluated on all 100 episodes and reached $\mathrm{SR}\geq50\%$. The frontier is the best score reached so far, and dashed orange segments mark drops between consecutive candidates.

These configurations trace the harness's development: front-view observation and primitive stepping score 66\% SR; VLA delegation scores 72\% with the 4B policy and 76\% with the 8B policy; and combining VLA execution with spatial tools reaches a development best of 81\%. An early place-graph interface scores 54\%, and the consolidated interface scores 77\%. One review found that the task prompt described local movement, although no movement tool was available; later revisions added that tool and replaced the stitched panorama with four separate 512-pixel views. The resulting tool and visual-prompt studies are reported in \Cref{tab:planner_models,fig:spatial_context}, and the resulting tool interface in \Cref{tab:tool_implementation}.

\subsection{Harnessing and VLA Ablation}
\label{sec:planner_models}

\Cref{tab:planner_models} measures how much each part of the harness contributes and what the VLA adds on top. Two language-model Planners, GPT-6 Astra and Claude Opus 5, navigate the same 100 R2R-CE episodes under the same episode limits and with the same 8B NavGPT VLA. Only the tools exposed to the Planner change between rows, and the task prompt describes only those tools. The tools and their visual returns were developed through RSI (\Cref{app:tool_development}).

\begin{table}[t]
\centering
\caption{\textbf{Tool and VLA ablation on the R2R-CE subset.} Resource use is averaged per episode. Opus costs are as reported by the provider's SDK; Astra costs are estimated from recorded usage at list prices. Inference excludes communication and harness overhead. Reaction is the mean interval between environment actions: Planner time per turn without the VLA, VLA inference per step with it.}
\label{tab:planner_models}
\label{tab:tool_ablation}
\vspace{3pt}
\tabstyle
\setlength{\tabcolsep}{1.0pt}
\begin{tabular*}{\linewidth}{@{\extracolsep{\fill}}lccccccccccccc@{}}
\toprule
\multirow{2}{*}{Planner} & \multicolumn{3}{c}{Tools} & \multicolumn{4}{c}{Navigation} & \multicolumn{5}{c}{Resource use} \\
\cmidrule(lr){2-4}\cmidrule(lr){5-8}\cmidrule(lr){9-13}
& VLA & Map & Backtrack & NE$\downarrow$ & OSR$\uparrow$ & SR$\uparrow$ & SPL$\uparrow$ & Tokens (k)$\downarrow$ & Turns$\downarrow$ & Cost (\$)$\downarrow$ & Infer. (s)$\downarrow$ & React. (s)$\downarrow$ \\
\midrule
VLA only & \checkmark & -- & -- & 4.23 & 78 & \mc{9}{71} & \mc{26}{66.14} & -- & -- & -- & \costc{4}{\textbf{44.1}} & \costc{4}{\textbf{0.78}} \\
\midrule
GPT-6 Astra & -- & -- & -- & 4.67 & 74 & \mc{11}{72} & \mc{21}{62.26} & \costc{12}{325.0} & \costc{21}{32.14} & \costc{14}{0.593} & \costc{8}{105.7} & \costc{8}{3.3} \\
GPT-6 Astra & -- & \checkmark & \checkmark & \textbf{2.55} & \textbf{87} & \mc{30}{\textbf{83}} & \mc{30}{69.04} & \costc{6}{190.8} & \costc{8}{10.71} & \costc{9}{0.469} & \costc{6}{66.8} & \costc{12}{6.2} \\
GPT-6 Astra & \checkmark & -- & -- & 4.58 & 79 & \mc{16}{75} & \mc{28}{67.83} & \costc{7}{227.9} & \costc{4}{\textbf{3.37}} & \costc{9}{0.469} & \costc{7}{84.4} & \costc{4}{\textbf{0.78}} \\
GPT-6 Astra & \checkmark & \checkmark & -- & 4.01 & 79 & \mc{18}{76} & \mc{28}{67.35} & \costc{6}{201.4} & \costc{4}{3.42} & \costc{8}{0.443} & \costc{6}{80.4} & \costc{4}{\textbf{0.78}} \\
GPT-6 Astra & \checkmark & \checkmark & \checkmark & 3.34 & 84 & \mc{28}{82} & \mc{30}{69.12} & \costc{4}{\textbf{155.0}} & \costc{4}{3.88} & \costc{6}{0.391} & \costc{6}{79.1} & \costc{4}{\textbf{0.78}} \\
\midrule
Opus 5 & -- & -- & -- & 5.43 & 70 & 66 & \mc{13}{55.24} & \costc{30}{721.0} & \costc{30}{47.93} & \costc{18}{0.69} & \costc{30}{425.9} & \costc{15}{8.9} \\
Opus 5 & -- & \checkmark & \checkmark & 3.45 & 85 & \mc{12}{73} & 44.82 & \costc{19}{483.4} & \costc{12}{16.75} & \costc{30}{0.990} & \costc{23}{323.4} & \costc{30}{19.3} \\
Opus 5 & \checkmark & -- & -- & 3.78 & 80 & \mc{18}{76} & \mc{21}{61.85} & \costc{6}{199.8} & \costc{9}{11.19} & \costc{4}{\textbf{0.353}} & \costc{15}{201.5} & \costc{4}{\textbf{0.78}} \\
Opus 5 & \checkmark & \checkmark & -- & 4.04 & 85 & \mc{26}{81} & \mc{30}{\textbf{69.29}} & \costc{7}{211.8} & \costc{7}{8.96} & \costc{6}{0.393} & \costc{15}{199.4} & \costc{4}{\textbf{0.78}} \\
Opus 5 & \checkmark & \checkmark & \checkmark & 3.69 & 83 & \mc{23}{79} & \mc{27}{66.70} & \costc{5}{185.6} & \costc{7}{8.50} & \costc{6}{0.397} & \costc{14}{191.8} & \costc{4}{\textbf{0.78}} \\
\bottomrule
\end{tabular*}
\par\vspace{2pt}
\end{table}

\paragraph{What the Planner can do in each configuration.}
The eight tools fall into four groups (\Cref{tab:tool_implementation}). Every configuration includes the four \emph{basic} tools: \texttt{observe\_forward} and \texttt{observe\_panorama} to look at the current surroundings, \texttt{navigate\_relative} to turn and move by a given angle and distance, and \texttt{terminate\_episode} to end the episode. With the basic tools alone, the Planner moves the agent itself, so every environment action is a Planner decision. Three groups can be added. The \emph{VLA} group, \texttt{navigate\_by\_instruction}, hands the full instruction, or the part that remains, to NavGPT VLA, which drives the agent until it signals a stop or reaches the delegation limit; the Planner then receives up to eight keyframes of the executed route, the endpoint views, and the execution status. A VLA stop ends only the delegation, and the episode ends only when the Planner calls \texttt{terminate\_episode}. The \emph{Map} group, \texttt{observe\_map}, returns a top-down map of the area observed so far, on which visited places are numbered in visiting order. The \emph{Backtrack} group, \texttt{navigate\_to\_node} and \texttt{annotate\_node}, lets the Planner return the agent to any numbered place and label places with what it has recognized there, so the map becomes something it can act on rather than only read.

\begin{table*}[t]
\centering
\scriptsize
\setlength{\tabcolsep}{3pt}
\begin{tabular}{p{0.075\textwidth}p{0.205\textwidth}p{0.24\textwidth}p{0.40\textwidth}}
\toprule
\textbf{Group} & \textbf{Tool} & \textbf{Effect on state} & \textbf{Returned to the Planner} \\
\midrule
Basic & \texttt{observe\_forward()} & None & Current forward RGB for fine semantic inspection. \\
 & \texttt{observe\_panorama()} & Updates observed local coverage, not pose & Four labelled views with bearings, clearance, pose displacement, and node information. \\
 & \texttt{navigate\_relative()} & Executes a validated local turn and translation & Realized motion status, endpoint panorama, and nearby node candidates. \\
 & \texttt{terminate\_episode()} & Optionally returns to a selected node, then stops & Stop or return status; this call ends the episode and cannot be undone. \\
\midrule
VLA & \makecell[l]{\texttt{navigate\_by\_}\\\texttt{instruction()}} & Runs one VLA rollout; the VLA keeps its history & Up to eight sampled route keyframes, endpoint panorama, optional observed map, and execution status. \\
\midrule
Map & \texttt{observe\_map()} & Updates observed coverage, not pose & Observed-only top-down map, map status, and route-ordered node table. \\
\midrule
Backtrack & \texttt{navigate\_to\_node()} & Returns to a visited node or a waypoint of the latest rollout & Endpoint panorama, observed map, return status, and updated node table. \\
 & \texttt{annotate\_node()} & Attaches a label to an existing node & Annotation status and updated node table. \\
\bottomrule
\end{tabular}
\caption{Tools that implement the navigation capabilities, grouped as in the ablation of \Cref{tab:planner_models}.}
\label{tab:tool_implementation}
\end{table*}

\paragraph{Resource use.}
Each row reports per-episode averages. Tokens count all input, including cached input, and all output; turns count Planner responses separated by environment feedback; cost is the provider's charge; and inference time is the total model inference time, excluding communication and harness overhead. Reaction time is the mean interval between environment actions: the Planner's time per turn without the VLA, and the VLA's time per step with it (\Cref{app:experiment_protocols}).

\paragraph{The harness sets how well the language model navigates.}
\label{sec:tool_additions}
With only the basic tools and without specific harnesses, both Planners perform poorly and spend heavily: GPT-6 Astra reaches 72 SR at 325.0k tokens per episode and Claude Opus 5 reaches 66 SR at 721.0k (\Cref{tab:planner_models}). A more robust, higher-quality harness changes this. Mapping tools lets the Planner assemble a much stronger spatial context than the current view provides, and Backtrack lets it act on that context far more efficiently. Together they raise success to 83 and 73 SR while cutting tokens by 41\% and 33\%, and Astra's tools-only row is the most accurate in the table.

\paragraph{Whether the VLA helps depends on the language model.}
The VLA always shortens the reasoning loop, but it does not always improve without a good harness. For Opus it does: turns fall from 16.75 to 8.50--11.19, success rises from 73 to 76--81 SR, and cost falls from \$0.99 to \$0.35--0.40 per episode. For Astra it does not: turns fall from 10.71 to 3.37, yet success drops to 75 SR and cost stays at \$0.469, as the tokens saved on navigation are spent supervising and correcting the VLA's execution. A stronger language model thus absorbs more of the benefit the VLA would otherwise provide, and the upper bound on success is set by the language model itself: once it reads the environment well enough to decide accurately on its own, success is limited by its own navigation ability rather than by the interface. Once the VLA is given its own harness, with Mapping and Backtracking over its rollouts, token use reaches its minimum for both models (155.0k for Astra and 185.6k for Opus) with success close to each model's best.

\paragraph{The VLA sets the minimum reaction time.}
Without the VLA, the system acts only as often as the language model decides, once every 3--19\,s on average depending on the model and tools. With the VLA, the control loop runs at VLA inference rate instead, 0.78\,s per step on average for the 8B model, which gives the whole system a basic guarantee of real-time response (\Cref{tab:planner_models}). \Cref{sec:interruption} examines how much of this higher interaction rate the Planner can exploit by checking VLA execution at increasing frequencies.

\subsection{Visual Prompt Organization}
\label{sec:runtime_ablation}
\suppressfloats[t]

\begin{wrapfigure}{r}{3.45in}
\vspace{-12pt}
\centering
\includegraphics[width=3.45in]{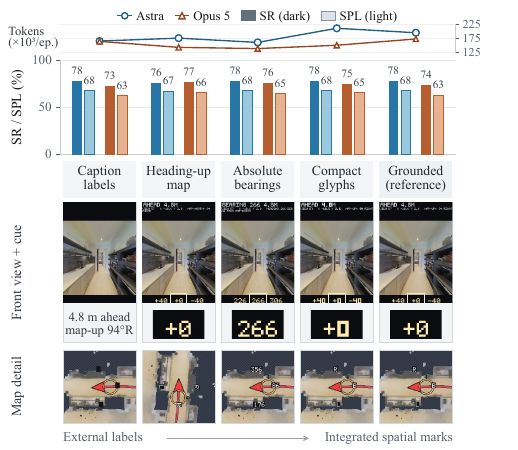}
\caption{\textbf{Visual prompt organization.} SR/SPL and token usage for Astra and Opus 5.}
\label{fig:spatial_context}
\vspace{-10pt}
\end{wrapfigure}

The harness also decides how tool returns are shown to the Planner: as a visual prompt of camera views and an observed map, annotated with view labels, node IDs, heading, and turn cues. \Cref{fig:spatial_context} varies one presentation factor at a time against this grounded reference while keeping tools fixed (\Cref{app:visual_prompt}): caption labels move the annotations into adjacent text, compact glyphs use a smaller digit font, heading-up rotates the map with the agent, and absolute bearings replace relative turn cues with fixed-frame directions.
Both language models are robust to how the visual prompt is organized: SR stays within 76--78\% for Astra and 73--77\% for Opus 5 across all five versions.

What changes is how much reasoning a version demands, and that demand does not follow human readability. Absolute bearings, the hardest version for a person to read, use the fewest tokens for both models, while the most expensive version differs between them: compact glyphs for Astra and the grounded reference for Opus, within 160--210k and 139--173k mean cumulative tokens per episode, respectively. Strong language models are therefore robust to visual context, and what they read easily does not follow human readability.

\begin{table}[tb]
\centering
\caption{\textbf{Check frequency.} VLA steps per check.}
\label{tab:interrupt}
\tabstyle
\footnotesize
\setlength{\tabcolsep}{4pt}
\begin{tabular}{lcccccc}
\toprule
Frequency & NE$\downarrow$ & OSR$\uparrow$ & nDTW$\uparrow$ & SR$\uparrow$ & SPL$\uparrow$ & Tokens (k)$\downarrow$ \\
\midrule
None & 3.82 & 81 & 16.74 & 77 & 67.51 & \textbf{179.2} \\
32 & 3.56 & 81 & 16.66 & 78 & 68.88 & 1,529.9 \\
16 & \textbf{2.01} & \textbf{85} & \textbf{18.72} & \textbf{83} & \textbf{77.36} & 2,319.5 \\
\bottomrule
\end{tabular}
\end{table}

\subsection{Communication Frequency and Cost}
\label{sec:interruption}

\Cref{tab:interrupt} varies how often the Planner checks VLA execution, measured in VLA forward passes per check; each pass predicts eight waypoints and executes the first (\Cref{sec:vla_policy}). Without periodic checks, the Planner regains control only when the VLA stops or reaches its delegation limit, after a median of 51 steps (90th percentile 123). At each check, the Planner reviews the route so far while the VLA keeps executing, and interrupts it only to revise or terminate the route. We test GPT-6 Astra with shared tools and instructions.

More frequent checks improve the route, but only at a steep cost. Checking every 32 steps gives most episodes at most one check and barely changes results (78 vs.\ 77 SR). Checking every 16 steps, about three times per episode, raises SR to 83 and SPL from 67.51 to 77.36, and cuts NE from 3.82 to 2.01\,m. Tokens grow much faster than accuracy, because each check re-reads the accumulated route context: they rise from 179.2k to 1,529.9k per episode at 32 steps, and to 2,319.5k at 16. Periodic checks thus buy accuracy with costly supervision, which motivates event-triggered review for real-world deployment.

\subsection{Overall Navigation Performance}
\label{sec:system_results}

\begin{table*}[t]
  \centering
  \footnotesize
  \caption{\textbf{Vision-and-language navigation.} Evaluation results on full R2R-CE and RxR-CE Val-Unseen splits. $^\dagger$RxR Val-Unseen English human followers \citep{anderson2020rxr}, measured in DE. Both \navgpt systems use the 8B VLA.}
  \label{tab:sota}
  \setlength{\tabcolsep}{2pt}
  \renewcommand{\arraystretch}{1.10}
  \begin{tabular*}{\linewidth}{@{\extracolsep{\fill}}llcccccccc@{}}
    \toprule
    \multirow{2}{*}{\textbf{Method}} & \multirow{2}{*}{\textbf{Model}} & \multicolumn{4}{c}{\textbf{R2R-CE Val-Unseen}} & \multicolumn{4}{c}{\textbf{RxR-CE Val-Unseen}} \\
    \cmidrule(lr){3-6} \cmidrule(lr){7-10}
    & & NE$\downarrow$ & OSR$\uparrow$ & SR$\uparrow$ & SPL$\uparrow$ & NE$\downarrow$ & nDTW$\uparrow$ & SR$\uparrow$ & SPL$\uparrow$ \\
    \midrule
    Human$^\dagger$~\citep{anderson2020rxr} & -- & -- & -- & -- & -- & 1.32 & 77.7 & 90.4 & -- \\
    \midrule
    NavFoM~\citep{zhang2025embodied} & 7B & 4.61 & 72.1 & 61.7 & 55.3 & 4.74 & 65.8 & 64.4 & 56.2 \\
    ABot-N0~\citep{abotN0} & 4B & 3.78 & 70.8 & \mc{7}{66.4} & \mc{14}{63.9} & 3.83 & -- & \mc{6}{69.3} & \mc{6}{60.0} \\
    AstraNav-World~\citep{astranavworld} & 3B+5B & 3.86 & 73.9 & \mc{9}{67.9} & \mc{16}{65.4} & 3.82 & -- & \mc{10}{72.9} & \mc{9}{61.5} \\
    OmniNav~\citep{omninav} & 3B & 3.74 & 74.6 & \mc{12}{69.5} & \mc{17}{66.1} & 3.77 & -- & \mc{11}{73.6} & \mc{10}{62.0} \\
    Qwen-RobotNav~\citep{qwennav} & 4B & 3.80 & 77.2 & \mc{12}{69.5} & \mc{13}{63.6} & 3.80 & 71.9 & \mc{12}{75.2} & \mc{15}{65.0} \\
    Qwen-RobotNav~\citep{qwennav} & 8B & 3.53 & 78.5 & \mc{16}{72.1} & \mc{18}{66.6} & 3.58 & 72.5 & \mc{14}{76.5} & \mc{16}{65.7} \\
    Robostral Navigate~\citep{robostralnavigate} & 8B & 3.20 & 81.3 & \mc{24}{77.4} & \mc{30}{\textbf{74.2}} & 3.47 & -- & \mc{12}{75.1} & \mc{21}{68.7} \\
    \midrule
    \multicolumn{10}{l}{\textit{Ours: executable navigation policy}} \\
    NavGPT VLA & 4B & 3.32 & 81.67 & \mc{19}{74.33} & \mc{20}{68.13} & 2.87 & 75.79 & \mc{20}{81.41} & \mc{23}{69.86} \\
    NavGPT VLA & 8B & 2.97 & 82.49 & \mc{23}{76.94} & \mc{24}{70.61} & 2.65 & 76.74 & \mc{21}{82.39} & \mc{24}{70.56} \\
    \midrule
    \multicolumn{10}{l}{\textit{Ours: complete navigation system}} \\
    \textbf{\navgpt} & Opus 5 & 2.77 & 84.07 & \mc{26}{79.01} & \mc{24}{70.16} & 2.27 & 75.09 & \mc{24}{84.80} & \mc{23}{69.85} \\
    \textbf{\navgpt} & GPT-6 Astra & \textbf{2.18} & \textbf{86.99} & \mc{30}{\textbf{81.51}} & \mc{24}{70.42} & \textbf{1.56} & \textbf{78.47} & \mc{30}{\textbf{90.43}} & \mc{30}{\textbf{74.31}} \\
    \bottomrule
  \end{tabular*}
\end{table*}

\begin{table}[t]
\centering
\begin{minipage}[b]{0.37\linewidth}
\caption{\textbf{Object-goal navigation.} Prior methods use HM3D v1, others HM3D v2. $\dagger$: \mbox{500-episode} subsets, scored with a stricter stop criterion. $\ddagger$: depth or \mbox{odometry}.}
\label{tab:objectnav}
\end{minipage}\hfill
\begin{minipage}[b]{0.61\linewidth}
\caption{\textbf{Embodied question answering.} Accuracy and normalized steps~\citep{zhai2025memoryeqa} on HM-EQA~\citep{ren2024explore} and MT-HM3D~\citep{zhai2025memoryeqa}; LLM score and path efficiency on EXPRESS-Bench~\citep{jiang2025express}. $^*$: from prior work. \navgpt uses a GPT-6 Astra judge.}
\label{tab:eqa}
\end{minipage}
\par\vspace{3pt}\nointerlineskip
\begin{minipage}[t]{0.37\linewidth}
\tabstyle\scriptsize
\setlength{\tabcolsep}{1pt}
\renewcommand{\arraystretch}{1.262}
\begin{tabular*}{\linewidth}[t]{@{\extracolsep{\fill}}lcccc@{}}
\toprule
\multirow{2}{*}{Model} & \multicolumn{2}{c}{MP3D} & \multicolumn{2}{c}{HM3D} \\
\cmidrule(lr){2-3}\cmidrule(lr){4-5}
& SR$\uparrow$ & SPL$\uparrow$ & SR$\uparrow$ & SPL$\uparrow$ \\
\midrule
\multicolumn{5}{l}{\textit{Prior methods: HM3D v1}} \\
VLFM$^{\ddagger}$ & 36.4 & \mc{5}{17.5} & 52.5 & \mc{10}{30.4} \\
SG-Nav$^{\ddagger}$ & \mc{5}{40.2} & 16.0 & \mc{2}{54.0} & 24.9 \\
WMNav$^{\ddagger}$ & \mc{12}{45.4} & \mc{4}{17.2} & \mc{7}{58.1} & \mc{12}{31.2} \\
CogNav$^{\ddagger}$ & \mc{13}{46.6} & 16.1 & \mc{26}{72.5} & \mc{2}{26.2} \\
Uni-NaVid & -- & -- & \mc{28}{73.7} & \mc{23}{37.1} \\
\midrule
Qwen-RobotNav 4B & \mc{21}{52.2} & 16.0 & \mc{30}{75.6} & \mc{11}{30.6} \\
Qwen-RobotNav 8B & \mc{16}{48.8} & \mc{6}{17.7} & \mc{24}{71.2} & \mc{15}{33.0} \\
\midrule
{\navgpt}$^{\dagger}$ & \mc{30}{59.2} & \mc{30}{25.2} & \mc{30}{75.3} & \mc{30}{40.8} \\
\bottomrule
\end{tabular*}
\end{minipage}\hfill
\begin{minipage}[t]{0.61\linewidth}
\tabstyle\scriptsize
\setlength{\tabcolsep}{1pt}
\renewcommand{\arraystretch}{1.155}
\begin{tabular*}{\linewidth}[t]{@{\extracolsep{\fill}}lcccccc@{}}
\toprule
\multirow{2}{*}{System} & \multicolumn{2}{c}{HM-EQA} & \multicolumn{2}{c}{MT-HM3D} & \multicolumn{2}{c}{EXPRESS} \\
\cmidrule(lr){2-3}\cmidrule(lr){4-5}\cmidrule(lr){6-7}
& Acc.$\uparrow$ & Steps$\downarrow$ & Acc.$\uparrow$ & Steps$\downarrow$ & Score$\uparrow$ & Path eff.$\uparrow$ \\
\midrule
\multicolumn{7}{l}{\textit{Prior systems}} \\
Explore-EQA~\citep{ren2024explore} & \mc{9}{58.4} & 0.52 & 36.2$^*$ & 0.64 & -- & -- \\
Graph-EQA~\citep{saxena2024grapheqa} & \mc{15}{63.5} & 0.20 & \mc{15}{45.6$^*$} & 0.45 & -- & -- \\
Memory-EQA~\citep{zhai2025memoryeqa} & \mc{12}{61.4} & 0.40 & \mc{11}{43.1} & 0.41 & -- & -- \\
Fine-EQA~\citep{jiang2025express} & \mc{6}{56.0} & 0.54 & -- & -- & 63.95 & 25.58 \\
3D-Mem~\citep{yang2025threedmem} & 50.4 & 0.63 & -- & -- & -- & -- \\
FAST-EQA~\citep{zhang2026fasteqa} & \mc{21}{69.2} & 0.65 & \mc{22}{50.5} & 0.52 & \mc{9}{68.7} & 29.25 \\
\midrule
\multicolumn{7}{l}{\textit{Qwen-RobotNav 8B executor}} \\
NavMCP~\citep{lei2026navmcp} & \mc{29}{76.7} & 0.15 & \mc{29}{54.4} & 0.19 & \mc{28}{79.27} & 33.96 \\
\midrule
\navgpt & \mc{30}{\textbf{77.38}} & 0.16 & \mc{30}{\textbf{55.27}} & 0.18 & \mc{30}{\textbf{80.43}} & 34.26 \\
\bottomrule
\end{tabular*}
\end{minipage}
\end{table}

\Cref{tab:sota} compares the complete \navgpt system with published methods on the full R2R-CE and RxR-CE Val-Unseen splits. Without any harness, NavGPT VLA is already at the state of the art on RxR-CE. The runtime then raises R2R-CE success from 76.94 to 81.51 SR with GPT-6 Astra, and gains more on RxR-CE, where success reaches 84.80 and 90.43 SR against 82.39 for the specialist alone and nDTW improves to 78.47: RxR instructions are longer and specify the route in far more detail, which gives the returned route evidence, returns to recorded places, and local corrections more to act on. \navgpt matches RxR human followers in success (90.43 vs.\ 90.4 SR) and path fidelity (78.47 vs.\ 77.7 nDTW), the first autonomous agent to reach human-level performance on this benchmark~\citep{anderson2020rxr}.

\paragraph{Object navigation and embodied question answering.}
\Cref{tab:objectnav} evaluates \navgpt on 500-episode subsets of the MP3D and HM3D v2 validation splits against published full-split results, scoring success only when the agent stops within the distance threshold while facing the target. Both cross-task studies use the 8B NavGPT VLA under a Claude Opus 5 Planner. The two benchmarks locate the runtime's contribution differently: MP3D spans 21 goal categories, many of them small or fine-grained (cushion, towel, picture), and the gain appears in success, whereas HM3D v2 uses six large, visually distinctive categories on which published systems already succeed in roughly three of four episodes, and the gain appears in path efficiency. On embodied question answering (\Cref{tab:eqa}), \navgpt is level with the strongest prior system, Qwen-RobotNav~\citep{lei2026navmcp}: accuracy is ahead on both datasets at comparable normalized steps (\Cref{app:cross_task}).

\subsection{Codec across Training Scopes}
\label{app:codec_ablation}
\begin{figure}[t]
\centering
\includegraphics[width=\linewidth]{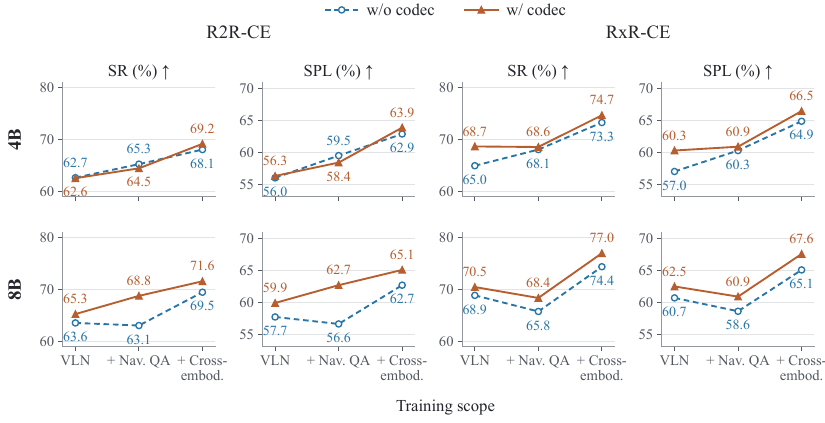}
\caption{\textbf{Codec comparison across training scopes.}}
\label{fig:codec_comparison}
\end{figure}

\Cref{fig:codec_comparison} compares \emph{Without codec}, the positional allocation of our reimplementation of the Qwen-RobotNav observation interface~\citep{qwennav}, with \emph{With codec}, our change-aware allocation, for 4B and 8B policies trained on three cumulative scopes, labelled VLN, + Nav.\ QA (adding embodied navigation question answering), and + Cross-embod.\ (further adding cross-embodiment navigation). All policies are evaluated standalone on the full R2R-CE and RxR-CE Val-Unseen splits.

Codec allocation helps in every 8B setting and on RxR-CE at both sizes. At 8B it raises R2R-CE SR by 1.7, 5.7, and 2.1 points across these three scopes, respectively, and RxR-CE SR by 1.6, 2.6, and 2.6 points; the large gain after adding navigation QA partly reflects a weaker positional baseline on that scope. At 4B it raises RxR-CE SR by 3.7, 0.5, and 1.4 points, whereas on R2R-CE it leaves SR within one point on the first two scopes ($-0.1$ and $-0.8$) and improves it only with cross-embodiment data ($+1.1$). With cross-embodiment data, codec allocation improves both SR and SPL on both benchmarks at both model sizes.

We see three likely reasons for this pattern. First, codec allocation matters most when images compete for a binding budget ($\rho_{\mathrm{vis}}<1$, \Cref{sec:vla_policy}). RxR-CE routes are longer and their instructions describe more landmarks along the way, so more steps run with a full history, and keeping detail on the views that change pays off; this is consistent with the gains on RxR-CE at both sizes. Second, the value of a change-aware rule grows with the visual diversity of the training data. In VLN data, every episode comes from one simulator with four fixed cameras and one motion scale, so how much a frame changes is largely predictable from its time and camera, and the positional rule is already close to the codec rule; Adding navigation QA brings question answering on the same episodes and therefore no new visual conditions. Cross-embodiment training adds single-camera views, slow and fast motion, moving targets, randomized camera heights and fields of view, and outdoor driving, where the amount of change between frames differs widely across sources. A fixed positional rule cannot adapt to these differences, whereas codec allocation follows the change in each sample. Third, codec allocation presents images at varying resolutions, which the model must learn to read. The 8B model exploits this on every scope, whereas the 4B model gains on R2R-CE only once cross-embodiment training supplies enough varied visual data.

\subsection{Model and Data Scaling}
\label{app:vla_scaling}

\begin{figure}[t]
\centering
\includegraphics[width=\linewidth]{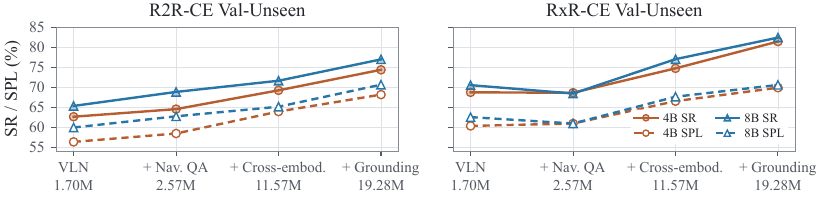}
\caption{\textbf{Model and data scaling.}}
\label{fig:scaling}
\end{figure}

\Cref{fig:scaling} plots 4B and 8B policies trained with codec across four cumulative training scopes on full Val-Unseen. Its first three scopes correspond to the \emph{With codec} curves in \Cref{fig:codec_comparison}; the fourth adds visual grounding and is the complete approximately 19.28M-record mixture used in the main comparison.

NavGPT VLA improves steadily with both training data and model size. Data contributes most: growing the mixture elevenfold, from 1.70M VLN records to the complete 19.28M, raises R2R-CE SR from 62.6 to 74.33 at 4B and from 65.3 to 76.94 at 8B, and RxR-CE SR from 68.7 to 81.41 and from 70.5 to 82.39. R2R-CE SR improves at every step; on RxR-CE, adding navigation QA leaves SR flat at 4B and lowers it at 8B (70.5 to 68.4), and the large gain arrives with cross-embodiment data. Model size adds a smaller but consistent margin: the 8B model leads the 4B model in R2R-CE SR at every scope, by 2.4 to 4.3 points. The data effect therefore outweighs doubling the parameters. With cross-embodiment data, the 4B model (69.2 SR) surpasses the 8B model trained on VLN only (65.3), and the complete 4B model (74.33) surpasses the 8B model trained without the final grounding step (71.6). The data curve has not saturated: the final step, which adds visual grounding and grows the mixture from 11.57M to 19.28M records, still raises R2R-CE SR by 5.1 and 5.3 points at 4B and 8B, and the 8B margin persists at the largest scale, suggesting further gains from more data and larger models. Because each step changes data volume and source composition together, these curves measure the cumulative recipe rather than the value of any single source.

\subsection{Real-World Deployment}
\label{app:deployment}
\label{app:device_validation}

We deploy \navgpt on a Unitree Go2 with RGB input, remote or onboard VLA inference, and a runtime LiDAR hazard monitor. In a matched comparison (\Cref{tab:device_results}), overlapping reasoning with priority monitoring raises SR from 73.3 to 83.3 over sequential queued handling and cuts p95 halt latency from 1840 to 180\,ms.

\subsubsection{Deployment Setup}

\paragraph{Robot and sensing.}
We use the Unitree Go2 platform and onboard VLA inference. NavGPT VLA receives the robot's front-facing RGB stream, with visual context allocated over the retained single-view history. This differs from the four-view simulator configuration in \Cref{sec:vla_policy}. The onboard LiDAR supplies geometric hazard detections to our runtime monitor; it is not an additional input to the VLA.

\paragraph{Remote and onboard inference.}
In the remote configuration, the robot compresses and transmits observations together with the Planner-selected navigation instruction to a server, which predicts waypoints and returns them for execution. The onboard configuration runs the 4B NavGPT VLA on an NVIDIA Jetson Thor with FP8 quantization and TensorRT acceleration, following \citet{qwennav}. Both configurations expose the same waypoint interface to the harness. Onboard inference removes the network round trip for VLA predictions; the hosted Planner remains a separate service. Model inference, communication, and physical execution contribute separately to end-to-end response time and are measured separately below.

\paragraph{Harness and runtime integration.}
The harness manages observations, spatial state, tool calls, and execution returns, while the runtime above it coordinates independently paced reasoning, execution, and monitoring. Moving the VLA between the server and the robot does not change this division of responsibilities. On the robot, every task-motion command passes through a final command filter, which forwards it only while the issuing operation holds the runtime's permission to move. The robot evaluation tests whether a LiDAR event can revoke that permission without waiting for the Planner, and whether execution resumes only after fresh sensing and a new permission (\Cref{app:priority_policy}).

\subsubsection{Glass-Door Execution Visualization}
\label{app:real_execution}

\begin{figure}[ht]
\centering
\includegraphics[width=\textwidth]{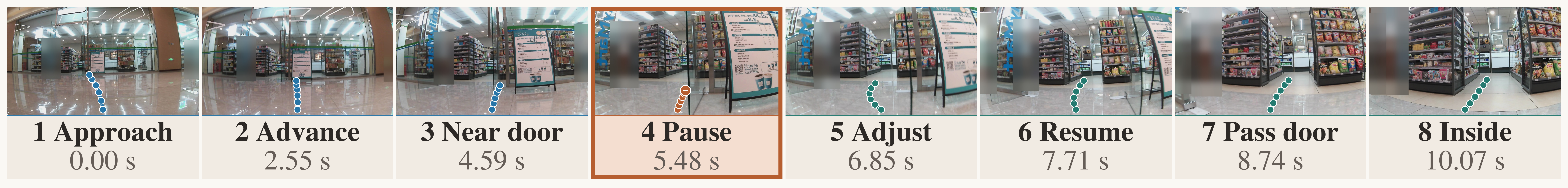}
\caption{Recorded glass-door execution. The outlined fourth panel highlights the near-door forward pause, followed by lateral correction and continued passage.}
\label{fig:real_execution}
\end{figure}

\paragraph{Approach and pause.}
\Cref{fig:real_execution} follows a recorded instruction to enter a shop. Panels~1--3 show the approach from an indoor concourse toward the already-open glass doorway, beside a freestanding sign. Panel~4 (5.48\,s) highlights the interruption of forward progress near the glass. Sensor poses place the pause and correction between 5.48 and 7.36\,s; lateral movement continues, so the robot is not completely stationary.

\paragraph{Correction and continuation.}
Panel~5 (6.85\,s) shows the lateral adjustment. Forward motion resumes in panel~6 (7.71\,s), the robot passes the entrance in panel~7, and panel~8 shows it inside the shop without visible contact with the glass door. The connected points illustrate an interrupted approach followed by revised paths through the opening: the task remains to enter the shop while the immediate route changes.

\paragraph{Connection to runtime control.}
This sequence illustrates the pause--revise--continue pattern targeted by the runtime: suspend the active motion operation, inspect fresh observations, and resume with an updated path under renewed authorization.

\subsubsection{Runtime Evaluation}

\paragraph{Controlled inputs.}
Conditions share the forward RGB stream, LiDAR detector and timestamped events, speed limit, routes, and manufacturer safeguards. The VLA placement and precision are held fixed within each runtime comparison, together with the single-camera input configuration, VLA checkpoint, Planner prompt, detector threshold, and firmware. An independent operator stop remains active; operator intervention counts as a failed trial.

\begin{table}[htbp]
\centering\small
\caption{Go2 scenarios and measurements.}
\label{tab:device_scenarios}
\setlength{\tabcolsep}{4pt}
\begin{tabular}{p{0.17\linewidth}p{0.38\linewidth}p{0.37\linewidth}}
\toprule
Scenario & Controlled event & Primary measurements \\
\midrule
A: Route revision & A detour or endpoint correction becomes necessary during an extended rollout; vary event timing relative to Planner inference. & Completion SR; event-to-authorized-repair time; route efficiency; Planner calls; execution idle time. \\
B: LiDAR protection & A soft obstacle enters a calibrated protected region during motion, including while the Planner is busy. Use matched harmless passages as negatives. & Missed and false stops; trigger-to-halt p50/p95; post-trigger distance; minimum clearance; operator interventions. \\
C: Priority and recovery & Trigger a hazard during repair or near VLA completion; delay a model reply, resend an outdated command to the robot's command filter, then clear the hazard. & Accepted stale commands; event ordering; halt latency; unauthorized resumption; correct resumption and task completion. \\
\bottomrule
\end{tabular}
\end{table}

\paragraph{Comparisons.}
We compare VLA-only and Planner-with-tools baselines with a matched $2\times2$ runtime comparison: sequential versus overlapping Planner--VLA execution, crossed with queued versus priority monitor handling. All four runtime configurations share the same models, tool definitions, route evidence, and LiDAR detections. Queued handling processes a detected event at the next regular control step; priority handling immediately revokes the active permission at the command filter. Keeping a priority monitor in the sequential condition tests whether protection can be explained by a simple independent stop gate, while the overlap contrast tests progress during reasoning. The final runtime configuration combines overlap with priority handling. These controls separate detector quality, scheduling, and authority transfer.

\begin{table}[htbp]
\centering\small
\caption{Go2 runtime results.}
\label{tab:device_results}
\setlength{\tabcolsep}{3pt}
\begin{tabular*}{\linewidth}{@{\extracolsep{\fill}}lcccccc@{}}
\toprule
& \multicolumn{2}{c}{A: Revision} & \multicolumn{2}{c}{B: Protection} & \multicolumn{2}{c}{C: Recovery} \\
\cmidrule(lr){2-3}\cmidrule(lr){4-5}\cmidrule(lr){6-7}
Condition & SR & Repair (s) & Missed/$N$ & Halt p95 (ms) & Stale/$N$ & Resume SR \\
\midrule
VLA-only & 60.0 & -- & -- & -- & -- & -- \\
Planner + tools & 70.0 & 14.2 & 9/30 & 2150 & 11/30 & 53.3 \\
Sequential + queued & 73.3 & 12.8 & 7/30 & 1840 & 8/30 & 60.0 \\
Sequential + priority & 76.7 & 11.5 & 2/30 & 420 & 3/30 & 70.0 \\
Overlapping + queued & 76.7 & 9.6 & 5/30 & 1310 & 6/30 & 66.7 \\
Overlapping + priority & 83.3 & 7.9 & 0/30 & 180 & 1/30 & 80.0 \\
\bottomrule
\end{tabular*}
\end{table}

\paragraph{Results.}
Overlap and priority contribute separately (\Cref{tab:device_results}). Overlapping Planner reasoning with VLA execution shortens repair from 12.8 to 9.6\,s under queued handling and from 11.5 to 7.9\,s under priority handling, because execution continues while the Planner reviews the route. Priority handling reduces missed stops from 7/30 to 2/30 in the sequential runtime and from 5/30 to 0/30 with overlap, and lowers p95 halt latency from 1840 to 420\,ms and from 1310 to 180\,ms, respectively. Combining both yields the highest revision SR (83.3 vs.\ 73.3 for sequential queued handling), the fewest accepted stale commands (1/30), and the highest resume SR (80.0).

\paragraph{Event timing and outcomes.}
We log observation compression, network transfer, and VLA inference durations separately, and record sensor capture, detector trigger, review start and end, permission revocation, the last accepted nonzero command, stop-command dispatch, physical halt, renewed permission, and resumed motion on a single synchronized clock. A stop command accepted by the robot's software interface does not by itself count as a halt: the robot has halted only when independently measured speed stays below a threshold, fixed before the trials, for a fixed interval. A stale command is one issued under a revoked permission version; Stale/$N$ counts trials in which such a command was accepted. Resume success requires fresh sensing, a new permission, and completion of the remaining task without operator intervention.

\paragraph{Trial allocation and uncertainty.}
Five pilot trials per condition and scenario validated logging and are excluded from the comparison. The evaluation uses five routes, three event timings, and two repetitions per condition and scenario (30 trials), randomized in matched blocks. Detection false negatives and false positives are annotated independently of the detector's output.

\subsection{Qualitative Results}
\label{app:qualitative}

\begin{figure}[p]
\centering
\includegraphics[width=\linewidth]{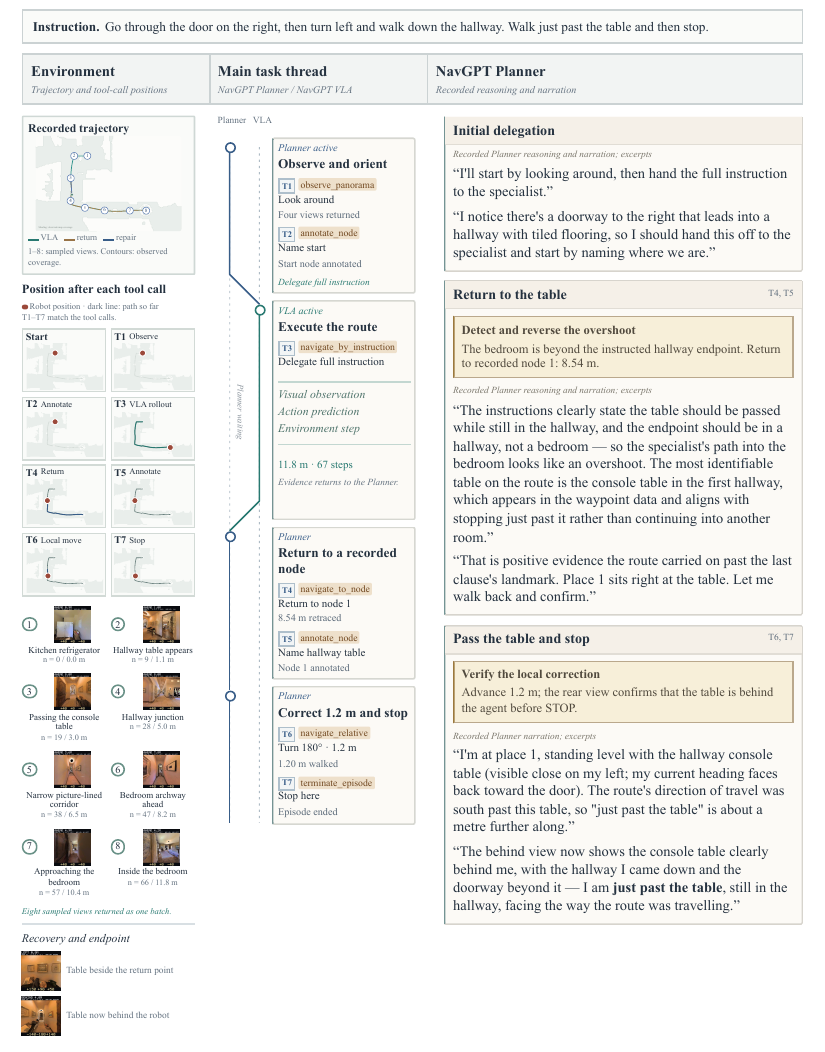}
\caption{\textbf{Backtracking to an overshot endpoint.} Claude Opus 5 on R2R-CE, episode 144. The Planner identifies the hallway table before the VLA's bedroom stop, retraces 8.54\,m to node 1, and verifies a 1.2\,m local correction before stopping.}
\label{fig:qualitative_backtrace}
\label{app:qualitative_backtrace}
\end{figure}

\begin{figure}[p]
\centering
\includegraphics[width=\linewidth]{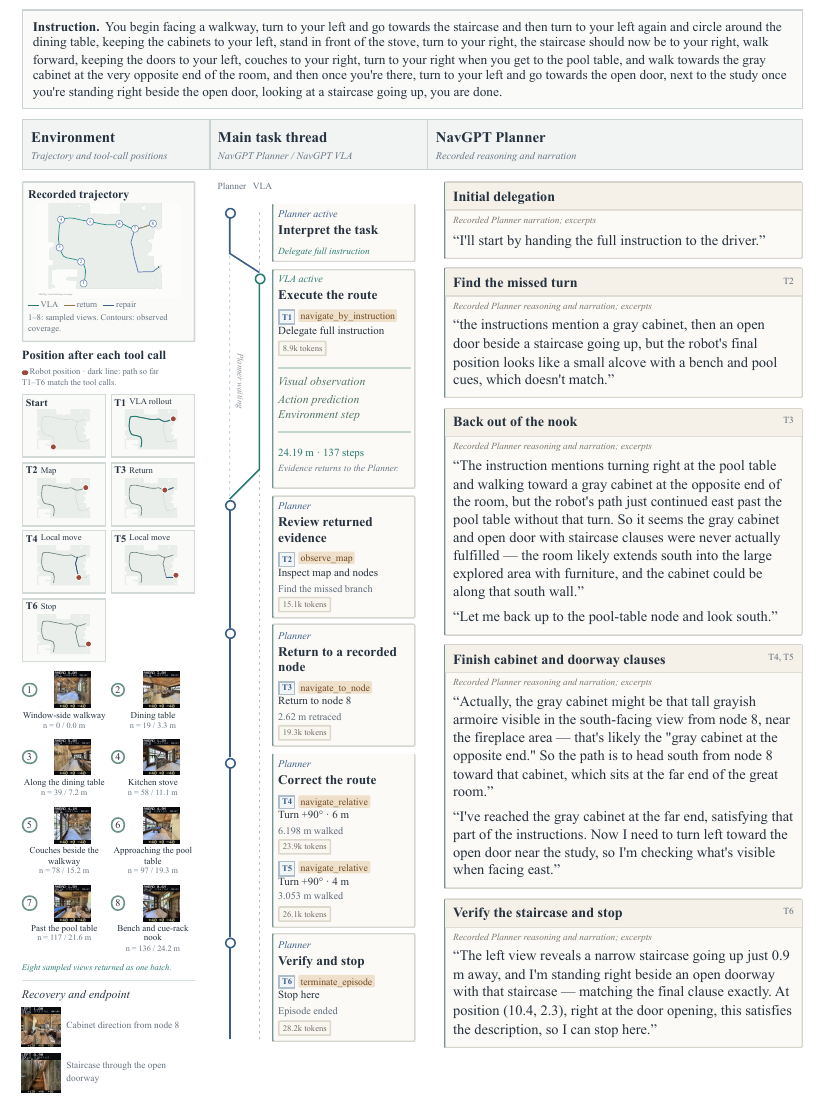}
\caption{\textbf{Completing omitted route clauses.} Claude Opus 5 on RxR-CE, episode 2523. The Planner returns from a dead-end nook to node 8, then uses relative motion to reach the gray cabinet and the open doorway with a rising staircase.}
\label{fig:qualitative_direct}
\label{app:qualitative_direct}
\end{figure}

\begin{figure}[p]
\centering
\includegraphics[width=\linewidth]{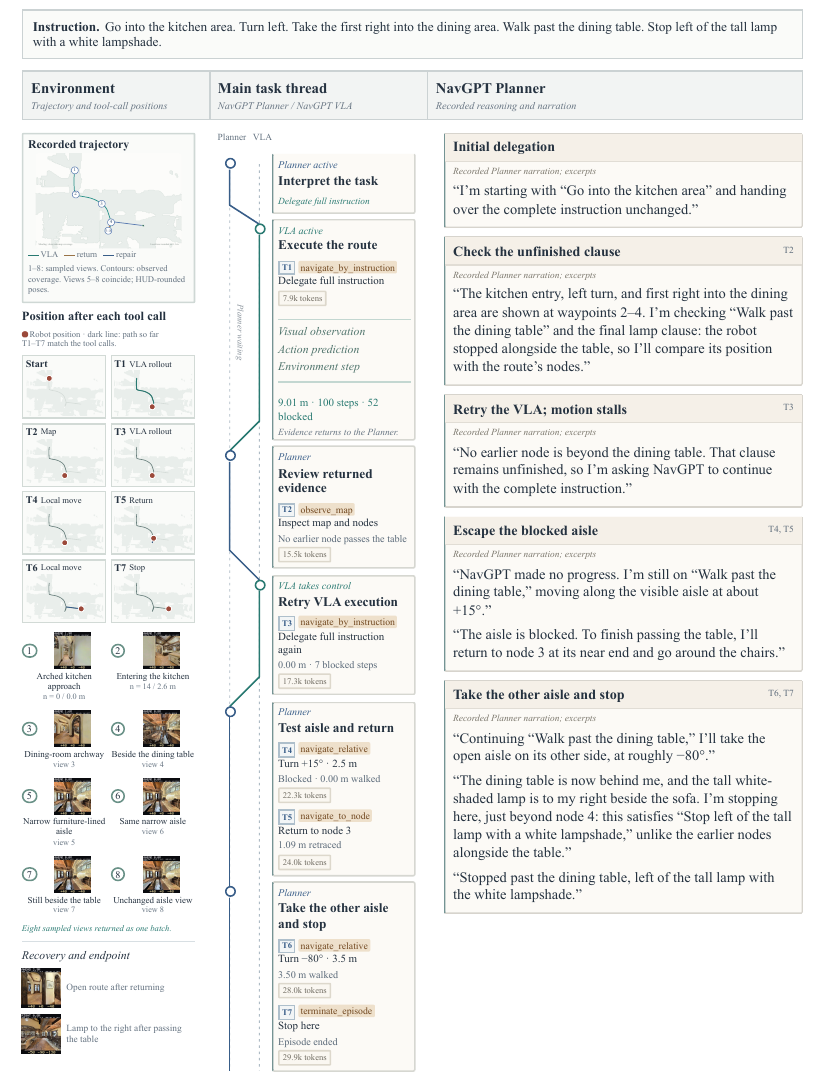}
\caption{\textbf{Recovering from a blocked aisle.} GPT-6 Astra on R2R-CE, episode 43. After repeated VLA execution and a direct motion attempt stall, the Planner returns to node 3, traverses the other aisle, and stops beyond the table beside the lamp.}
\label{fig:qualitative_blocked}
\label{app:qualitative_blocked}
\end{figure}

\begin{figure}[p]
\centering
\includegraphics[width=\linewidth]{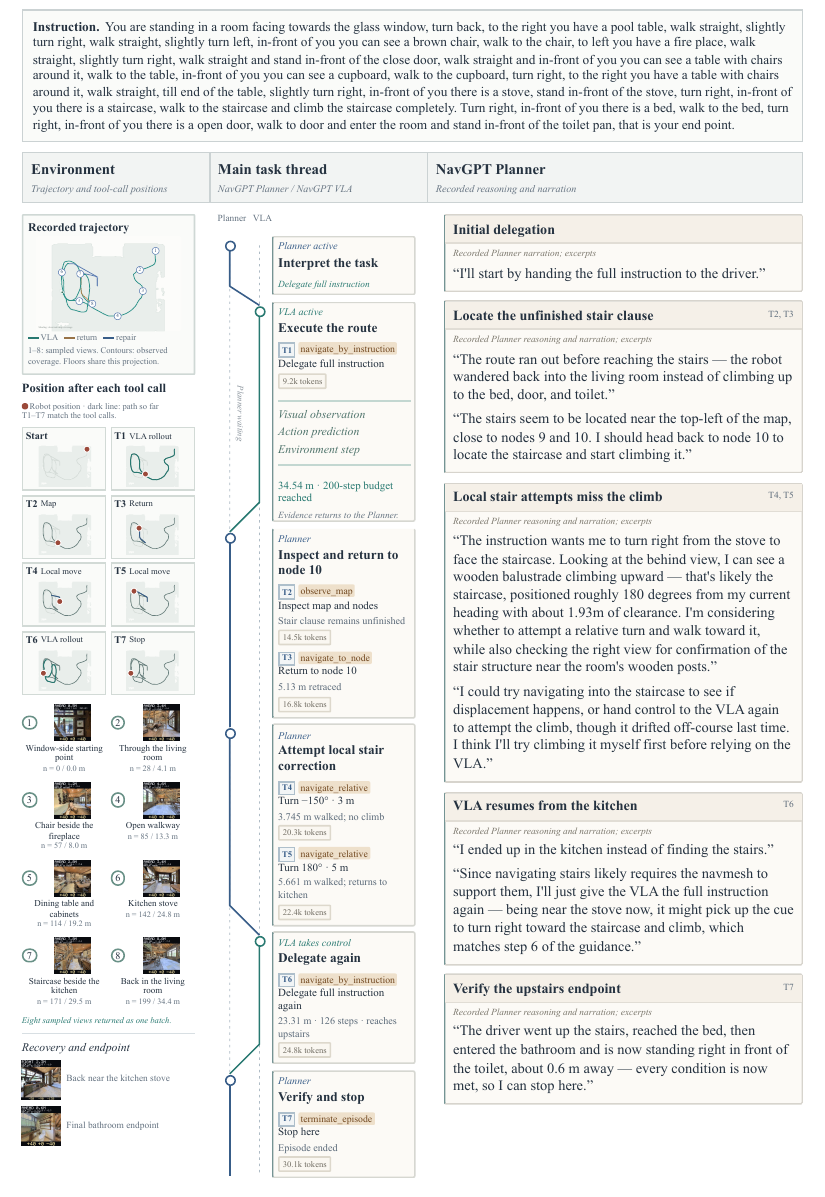}
\caption{\textbf{Resuming delegation after unsuccessful local correction.} Claude Opus 5 on RxR-CE, episode 5782. With the stair clause still unfinished, the Planner delegates the full instruction again. The second rollout reaches the upstairs bathroom, which the Planner verifies before stopping.}
\label{fig:qualitative_redelegate}
\label{app:qualitative_redelegate}
\end{figure}

Four selected successful episodes illustrate how returned observations guide the Planner after extended VLA execution. Claude Opus 5 recognizes that the VLA has passed the hallway table, retraces 8.54\,m to node 1, and verifies a 1.2\,m endpoint correction (\Cref{fig:qualitative_backtrace}). On a longer route, it returns from a dead-end nook and directly completes the omitted cabinet and doorway clauses (\Cref{fig:qualitative_direct}). GPT-6 Astra recovers from stalled execution: after a repeated VLA call and a blocked local move, it retreats to node 3 and takes the other aisle (\Cref{fig:qualitative_blocked}). Finally, Opus delegates again when local stair attempts leave the instruction unfinished; the next rollout reaches the upstairs bathroom (\Cref{fig:qualitative_redelegate}).

Each case includes its complete instruction. Maps T1 onward show the position after each tool call returns; numbered views are sampled from the first VLA rollout. Solid lines mark the thread that holds motion authority, and dashed lines mark waiting. Token counts include cached input and output.

\clearpage

\section{Conclusion}
\label{sec:conclusion}

We presented \navgpt, a complete navigation harness with a runtime layer above it. The harness organizes context and navigation state, exposes spatial tools, and returns execution evidence for repairing the route and deciding when the task is complete; the runtime coordinates these capabilities through thread control and motion authority. The standalone VLA and complete system attain strong full-split navigation results, with the latter matching human followers on RxR-CE under continuous rather than graph-based control. The harness studies show how spatial tools support Planner navigation, visual presentation affects processing cost, and VLA delegation provides frequent action updates with fewer reasoning turns. Together, these components establish a complete embodied interface through which reasoning can inspect, revise, and control extended physical execution. The harness provides the structure for integrating stronger reasoning and action models as they become available.

\clearpage
\subsection*{AI use statement}

Large language models are components of the studied system: GPT-6 Astra and Claude Opus 5 act as Planners, and a GPT-6 Astra judge scores EXPRESS-Bench answers, as described in \Cref{sec:experiments} and \Cref{app:experiment_protocols}. Separately, we used generative AI coding assistants (Claude Code and Codex) to help implement the experiment and evaluation infrastructure, to create and revise scientific figures, and to draft, edit, and format parts of this paper. We have reviewed all AI-assisted work: AI-assisted code was tested before use, and all text, figures, and reported numbers were checked by the authors against the recorded experiment outputs. We take responsibility for the final content of this work, including text, claims or artifacts produced with the aid of generative AI.

\clearpage
\bibliography{iclr2027_conference}
\bibliographystyle{iclr2027_conference}

\clearpage
\appendix
\section{Navigation Interface Implementation}
\label{app:runtime_interfaces}

\Cref{sec:tools} describes the navigation capabilities without reference to specific tool names. The full \navgpt configuration provides them through eight tools (\Cref{tab:tool_implementation}). The harness checks each call's tool name and arguments before executing it, the environment executes the resulting motion, and the harness keeps the records of the route and of visited places (nodes). The Planner cannot add tools through an instruction or change the navigation state directly.

\subsection{Division of Responsibilities}
\label{app:runtime_layers}

\Cref{fig:layers} lists, for each part of the \navgpt system, the state it owns and the decisions it makes. This division is independent of the runtime's threads and of where each model is deployed. The main task thread calls Planner reasoning, VLA execution, and other tools as its current operation. The Planner's context is assembled from several sources: captured observations, the maintained route record, formatted tool results, and the conversation history kept by the model provider.

\begin{figure}[!htbp]
\centering
\small
\setlength{\tabcolsep}{5pt}
\renewcommand{\arraystretch}{1.10}
\begin{tabular}{p{0.30\textwidth}p{0.62\textwidth}}
\toprule
\textbf{Responsibility} & \textbf{Component and role} \\
\midrule
Thread coordination & Runtime: schedule threads, manage permissions, and control motion authority. \\
Navigation decisions & Planner: interpret evidence, choose tools, and decide when the goal is reached. \\
Context construction & Harness: select and render the images and state shown to the models. \\
Persistent navigation state & Harness: keep route references, nodes, and annotations. \\
Tool checking & Harness: validate requests and execute permitted tool calls. \\
Low-level navigation & VLA: predict waypoints from its visual history. \\
Physical execution & Environment: execute motion and report sensing and collisions. \\
\bottomrule
\end{tabular}
\caption{Division of responsibilities in the \navgpt system.}
\label{fig:layers}
\end{figure}

\subsection{Capabilities and Execution}

\begin{table}[htbp]
\centering
\small
\setlength{\tabcolsep}{4pt}
\renewcommand{\arraystretch}{1.12}
\begin{tabular}{p{0.19\textwidth}p{0.32\textwidth}p{0.41\textwidth}}
\toprule
\textbf{Capability} & \textbf{Navigation purpose} & \textbf{What it takes and returns} \\
\midrule
Local observation & Ground landmarks and inspect the current surroundings & Relate forward or panoramic appearance to the current viewing direction and pose. \\
Instruction delegation & Execute an extended route with the learned policy & Pass the requested instruction to the VLA, keep its history across calls, and return images of the executed route with its status. \\
Route inspection & Relate current progress to previously encountered places & Show the observed map and the ordered list of visited places, without marking unobserved space as free. \\
Spatial revision & Revisit a recorded place or correct a local endpoint error & Turn a reference to a visited place, or a relative motion request, into checked movement and report what was executed. \\
Semantic annotation & Retain an interpretation of a visited place & Attach a label to a recorded place without changing its recorded position. \\
Task termination & Accept the final navigation state & Make episode completion explicit and distinct from the end of a delegated rollout. \\
\bottomrule
\end{tabular}
\caption{Navigation capabilities and their interfaces.}
\label{tab:tools}
\end{table}

\paragraph{Execution and returned results.}
Instruction delegation runs as the main task thread's current operation: it passes the instruction chosen by the Planner to the VLA, which keeps its history across steps, and executes the VLA's predictions until the VLA signals a stop, the step limit for one delegation is reached, or the episode limit is reached. A VLA stop ends only the delegation, not the episode. The instruction may be the full task instruction or one for a subtask. The observed map distinguishes unobserved space from observed free space, and numbered nodes give stable targets for returning. Annotations attach names, captions, or instruction clauses to existing nodes without changing the recorded map. Because ending the episode may include a return to a selected node, any restriction on returning applies to that option as well as to the dedicated return tool. The full configuration has no primitive motion tools, no tool for recalling arbitrary past images, and no object detector.

\section{Runtime State and Timing}
\label{app:runtime_formulation}

This appendix expands \Cref{eq:planner_choice,eq:foreground_transfer,eq:foreground_eligibility}, separating the harness interfaces that the models see from the thread control performed by the runtime above them.

\subsection{Logical Threads and Shared State}

We use $j$ for a logical thread, $k$ for an invocation, and $t$ for wall-clock time. Shared system state $S(t)$ combines the instructions and spatial records kept by the harness, observations from the environment, and the runtime's records of threads and running operations. Each thread carries
\begin{equation}
    S^{\mathrm{thread}}_j(t)=\bigl(\mathrm{role}_j(t),H_j(t),\mathrm{status}_j(t),\mathcal{R}_j(t),\mathcal{W}_j(t)\bigr),
    \label{eq:thread_state}
\end{equation}
where $\mathrm{role}_j$ is its responsibility, $H_j$ its working history, $\mathrm{status}_j$ its lifecycle state, and $\mathcal{R}_j,\mathcal{W}_j$ its permitted shared-state reads and proposed writes. Shared spatial history records the route and places, whereas working history records a thread's own interactions. The runtime tracks the status of each tool call separately, so the main thread can ask the Planner to reassess while its VLA rollout continues or is paused; finishing a tool call does not mean that the task is complete.

The control-transfer rule $\sigma$ retains, transfers, or revokes motion authority in response to events. A routine transfer withdraws the previous operation's permission to move and waits until its current motion step has finished before permitting new motion. Thread contexts survive a handoff, but outdated commands do not become valid again when execution resumes.

\subsection{Synchronous Implementation}

In a sequential loop, at handoff time $t_k$, with $S_k=S(t_k)$, main-thread history $H_k$, and available tools $\mathcal{T}_k$, each tool-call cycle is
\begin{equation}
\begin{aligned}
    C_k &= \contextmap_{\mathrm{main}}(S_k,H_k,\mathcal{T}_k),\\
    a_k &\sim \pi_{\mathrm{Planner}}(\cdot\mid C_k,\mathcal{T}_k),\qquad
    (S_{k+1},E_k)=\operatorname{Execute}(S_k,a_k).
\end{aligned}
\label{eq:runtime}
\end{equation}
The harness constructs context $C_k$, validates call $a_k$, executes it if permitted, and returns updated state and evidence $E_k$. The runtime controls thread execution and motion authority; the environment executes physical motion. A long operation such as a VLA rollout may span many observations and motion steps between two Planner calls. VLA execution and motion chosen directly by the Planner use the same interface.

\subsection{Context Construction and State Updates}

For thread $j$ at its $k$th invocation, the general context and decision interface is
\begin{equation}
    C_{j,k}=\contextmap_j\!\left(S|_{\mathcal{R}_j},H_j,\mathcal{E}_{j,k},\mathcal{T}_j\right),\qquad
    a_{j,k}\sim\pi_{j,k}\!\left(\cdot\mid C_{j,k},\mathcal{T}_j\right),
    \label{eq:thread_context}
\end{equation}
evaluated when the invocation starts, where $\mathcal{E}_{j,k}$ contains the events delivered to the thread and $S|_{\mathcal{R}_j}$ restricts the visible state to permitted reads. The harness decides what information is gathered, what is kept, and where each item came from; $\contextmap_j$ assembles one model request from the state and tools permitted to thread $j$. Because the hosted model provider keeps the conversation history, replaying a request exactly without the provider would also require a complete record of each request. The harness maintains the navigation records, while the environment executes physical motion. The runtime accepts a proposed update only if it lies within $\mathcal{W}_j$ and the current task and delegation permissions, and the component that owns the target state checks its format and target.

\subsection{Motion Authority and Protective Control}

Let $G_o(t)\in\{0,1\}$ indicate whether physical operation $o$ holds the permission to move, and let $g(t)$ be a permission version number that increases whenever a permission is revoked or replaced. Only a permitted operation of the foreground thread may hold the permission (\Cref{eq:foreground_eligibility}), and each task command $u_o$ carries the version number under which it was issued. A command is eligible only if
\begin{equation}
    \sum_o G_o(t)\le1,\qquad
    \operatorname{eligible}(u_o,t)=G_o(t)\,\mathbf{1}[\operatorname{version}(u_o)=g(t)]\,\mathbf{1}[\operatorname{preconditions}(u_o,S(t))],
    \label{eq:motion_authority}
\end{equation}
so late model outputs or waypoints queued before a pause cannot regain authority. With a protective-stop flag $p_{\mathrm{stop}}(t)$ that stays set until it is explicitly cleared, and a low-latency filter $\operatorname{Protect}$ over fresh sensor observations $O_{\mathrm{phys}}(t)$, the final command filter on the robot outputs
\begin{equation}
    u_{\mathrm{robot}}(t)=
    \begin{cases}
        u_{\mathrm{stop}},&p_{\mathrm{stop}}(t)=1\ \text{or no task command is eligible},\\
        \operatorname{Protect}\bigl(O_{\mathrm{phys}}(t),u_{\mathrm{task}}(t)\bigr),&\text{otherwise}.
    \end{cases}
    \label{eq:protective_control}
\end{equation}
An obstacle detection or a control-loop timeout can set $p_{\mathrm{stop}}$ and revoke the current permission without waiting for the Planner; recovery requires the stop to be cleared, the state to be refreshed, and a new permission. Only the Planner decides that the goal has been reached; the protective path can block motion regardless of that decision.

\subsection{Scheduling and Interaction Frequency}
\label{app:scheduling}

Activities may run at different update periods and on different events, as in full-duplex and asynchronous interaction--thinking models~\citep{defossez2024moshi,huang2026duplexomni}. The runtime design allows the Planner to review a fixed copy of the current state while VLA execution continues, interrupting only when the review asks for revision or termination. One VLA step predicts eight waypoints and executes the first; discarded predictions and failed retries are not counted as steps. The benefit of concurrent review is measured on the Go2 (\Cref{app:device_validation}). With a constant step duration $P_{\mathrm{VLA}}$ and a check every $F$ steps, $F P_{\mathrm{VLA}}$ is the execution time between checks; total elapsed time also includes any time spent paused for review.

\subsection{Priority and Handoff Policy}
\label{app:priority_policy}

The runtime assigns priorities when activities are created, from highest to lowest: protective stops and control-loop timeouts, episode limits, completion of an operation, and route-review requests. Each event carries its operation, permission version, timestamp, and sequence number; events of equal priority are handled in arrival order. Outdated review or completion events are discarded, and repeated review requests for the same operation are merged into one pending request. A protective stop immediately revokes motion authority and increases the version number; reasoning may continue, but neither a late model response nor a queued waypoint can move the robot under the old permission. Clearing an alert alone does not resume motion: fresh observations and a new permission are required. Episode limits also revoke motion; otherwise, an operation that finishes on its own takes precedence over a scheduled review arriving at the same time. A review that lets the route continue keeps the VLA's history, whereas a correction waits for the current motion step to finish before a new operation is permitted.

This policy describes the full runtime design. The simulator implementation used for our ablations is simpler: it executes one motion operation at a time and implements the version checks, the precedence of operations that finish on their own, a single pending review request, and continuation that keeps the VLA's history. Preemption by an independent safety monitor and its physical timing are measured on the Go2 (\Cref{app:device_validation}).

\subsection{Interaction-Frequency Evaluation}
\label{app:interrupt_protocol}

The check schedule is defined in \Cref{app:scheduling}. All frequencies use the same Planner, VLA, prompt, tools, route-summary format, and resumption procedure, which keeps the VLA's history; a check may let the current route continue without choosing a recovery action. The execution loop also records a review request after two consecutive blocked moves; such requests are delivered at the next scheduled check or when the VLA stops on its own, rather than triggering an extra check, so the frequency comparison contains no event-triggered reviews. A request cannot change navigation state or end the episode, and the runtime revokes the VLA's permission to move before permitting replacement motion. Cost includes all model calls and waiting time; periodic checks are counted separately from returns that occur when the VLA stops on its own. \Cref{tab:interrupt_permissions} summarizes what each component may do at a handoff.

\begin{table}[htbp]
\centering
\small
\caption{What each component may do when execution is interrupted.}
\label{tab:interrupt_permissions}
\setlength{\tabcolsep}{3pt}
\renewcommand{\arraystretch}{1.12}
\begin{tabular}{>{\raggedright\arraybackslash}p{0.23\linewidth}>{\raggedright\arraybackslash}p{0.20\linewidth}>{\raggedright\arraybackslash}p{0.21\linewidth}>{\raggedright\arraybackslash}p{0.27\linewidth}}
\toprule
Operation & Planner & VLA & Enforced by \\
\midrule
Interrupt the running VLA rollout & May request, if permitted & Stops after the current step & Runtime: revokes the permission to move \\
Request an early Planner review & Receives it, if permitted & Keeps executing & Runtime: queues or discards the request \\
Resume or revise the route & Chooses an available tool & Executes a valid delegation & Runtime: grants one operation the permission to move \\
Change physical or route state & Cannot change it directly & Proposes waypoints & Environment: executes motion; harness: updates route records \\
End the episode & Decides explicitly to stop & Its stop ends only the delegation & Runtime: enforces the stop and episode limits \\
Share information between Planner and VLA & Sends structured instructions only & Returns structured status and route evidence only & Runtime: controls visibility; harness: builds context \\
\bottomrule
\end{tabular}
\end{table}

\section{Visual-Context Allocation Details}
\label{app:codec_protocol}

\paragraph{Baseline and codec rule.}
The positional baseline reimplements the Qwen-RobotNav observation interface~\citep{qwennav}: time and camera weights allocate a shared budget across retained images, and each image is resized to its assigned pixel budget. The time weight of the $t$th of $T$ retained observations, counted from the oldest, is $\exp(\gamma t/(T-1))$, so $\gamma\geq0$ sets how strongly recent observations are favored. Codec adds the RGB-change multiplier defined in \Cref{eq:codec_score}, retaining the same allocator and image-resizing path. The score measures visual change, including changes caused by camera motion and lighting; it is not a learned semantic-importance estimate or an optical-flow estimator.

\paragraph{Integer allocation.}
Flatten the retained history in time-major, front/right/back/left order. For $M=N_{\mathrm{hist}}N_{\mathrm{view}}$ images, initialize every allocation to $b_{\min}$, requiring $Mb_{\min}\leq B_{\mathrm{vis}}$. Distribute the remaining budget proportionally to the positive joint weights, take the floor of each share, cap additions at $b_{\max}$, and redistribute any overflow among uncapped images. Assign the final integer remainder in descending weight order, filling each image's remaining capacity before advancing. Ties between equal weights are broken by a fixed sort order. No tokens are assigned beyond an image's capacity: when $B_{\mathrm{vis}}\geq Mb_{\max}$ all images receive $b_{\max}$ and the excess budget is unused. Image resizing preserves aspect ratio and rounds dimensions to the encoder's patch/merge grid, so realized image-token counts can be below their allocated limits.

\paragraph{Evaluation configuration.}
At evaluation, the four-view VLA uses $B_{\mathrm{vis}}=3072$, $\gamma=2$, camera weights $(2,1,0.5,1)$ in front/right/back/left order, $b_{\min}=4$, and $b_{\max}=196$. It retains at most 16 observations, always including the first and current ones, with the remaining indices sampled uniformly without replacement from intermediate history and sorted chronologically. With these settings, short histories of at most three observations saturate the per-image caps; allocation becomes budget-limited from four observations onward. These evaluation values differ from the randomized training budgets and from the settings of the synthetic check below.

\begin{table}[t]
  \centering
  \small
  \caption{\textbf{Allocation saturation.} Deterministic synthetic change-score diagnostic.}
  \label{tab:saturation}
  \setlength{\tabcolsep}{5pt}
  \begin{tabular}{rrrrcrr}
    \toprule
    $B_{\mathrm{vis}}$ & $N_{\mathrm{hist}}$ & $N_{\mathrm{view}}$ & $b_{\max}$ & $\rho_{\mathrm{vis}}$ & \makecell{Images\\differing} & \makecell{Mean $|\Delta b|$\\(tokens)} \\
    \midrule
    3072 & 2 & 1 & 196 & 7.84 & 0\,/\,2 & 0.0 \\
    3072 & 8 & 1 & 196 & 1.96 & 0\,/\,8 & 0.0 \\
    3072 & 15 & 1 & 196 & 1.04 & 0\,/\,15 & 0.0 \\
    4096 & 4 & 4 & 256 & 1.00 & 0\,/\,16 & 0.0 \\
    \midrule
    3072 & 16 & 1 & 196 & 0.98 & 3\,/\,16 & 8.0 \\
    3072 & 8 & 4 & 196 & 0.49 & 30\,/\,32 & 47.3 \\
    1024 & 16 & 1 & 196 & 0.33 & 14\,/\,16 & 31.4 \\
    2048 & 8 & 4 & 196 & 0.33 & 31\,/\,32 & 18.9 \\
    1024 & 16 & 4 & 196 & 0.08 & 57\,/\,64 & 7.9 \\
    \bottomrule
  \end{tabular}
\end{table}

\Cref{tab:saturation} is a synthetic check of the allocator, using $b_{\min}=4$, $\gamma=2$, the camera weights above (weight one for a single view), and each row's budget and cap. Change scores are drawn uniformly from $[0.05,1)$ with a fixed random seed, and the first retained image of every view is given score one. The table reports how many images receive different allocations under the two rules and the mean per-image $|\Delta b|$, where $\Delta b=b^{\mathrm{codec}}-b^{\mathrm{pos}}$. All integer bounds and allocation totals hold, and the four saturated rows, where the two rules must agree, give identical allocations. We will release the score arrays, allocations, and allocator code. These are synthetic calculations, not navigation results; navigation performance is evaluated in \Cref{fig:codec_comparison}.

\section{Additional Experimental Details}
\label{app:experiment_protocols}

\subsection{Evaluation Settings}

\begin{table}[htbp]
\centering
\small
\caption{Harness evaluation settings.}
\label{tab:run_manifest}
\setlength{\tabcolsep}{3pt}
\renewcommand{\arraystretch}{1.12}
\begin{tabular}{p{0.35\linewidth}p{0.57\linewidth}}
\toprule
Setting & Value \\
\midrule
RGB resolution & 512 pixels \\
Camera order & Front, right, back, left \\
Returned route keyframes & At most 8, uniformly sampled along the rollout \\
Episode / delegation limit & 500 / 200 environment steps \\
Planner-call limit & 200 turns per episode \\
Episode timeout & 2,400\,s \\
\bottomrule
\end{tabular}
\end{table}

Benchmark goal distances and success labels are available only to the offline evaluator.

\subsection{Human Performance on RxR-CE}
\label{app:human_reference}

RxR asks an agent to follow a route instruction to its endpoint. Human demonstrations use discrete graph edges, which abstract away low-level movement and obstacle avoidance; RxR-CE realizes navigation through continuous motion~\citep{anderson2020rxr,krantz2020beyond}. Our 90.43 SR and 78.47 nDTW match English human followers (90.4 SR and 77.7 nDTW) despite this additional control burden on the same instructions.

\subsection{Cross-Task Evaluation}
\label{app:cross_task}

The object-goal subsets are drawn in proportion to target categories and scenes. Reference values follow Table~4 of \citet{qwennav}. We do not report per-category MP3D results, since proportional sampling leaves the rarest categories with about two episodes each. Question-answering metrics follow Table~7 of \citet{qwennav}; prior EXPRESS-Bench scores used the now-deprecated GPT-4o-mini judge.

\subsection{Planner Scaffolds and Accounting}

We replace each scaffold's default task prompt with the \navgpt navigation prompt, which describes the robot, the instruction, the tools, the movement budget, and the stopping rule, and we provide the corresponding navigation tools. The scaffolds' own task prompts are not used, and every reported Planner result is produced by the \navgpt harness running inside them. The scaffold remains responsible for assembling the conversation and managing its context window, including any automatic compression by the provider. The harness determines which observations, route evidence, and spatial state enter those messages, subject to the runtime's visibility permissions. Replacing the task prompt therefore does not make Codex and Claude Code identical: their system instructions, built-in tools, and formatting of tool results may still differ. Ablations within one Planner keep the scaffold and prompt fixed; comparisons across Planners describe each complete Planner--scaffold combination rather than ranking the underlying models in isolation.

Tokens (k) sum all processed input tokens, including cached input, and generated output tokens across requests, divided by 1,000. Turns count Planner responses separated by environment feedback, including the initial response and final acknowledgement; text, reasoning, and parallel tool calls within a response count once. Cost is the total hosted inference charge in US dollars per attempted episode, including retries and cache charges. VLA latency is reported separately rather than priced. Opus costs are as reported by the provider's SDK; Astra costs use recorded token usage at list prices. Inference timing excludes communication and harness overhead. The action-update interval in \Cref{tab:planner_models} is Planner time per turn without the VLA and VLA inference per step with it; it does not include the time for an interrupt to reach the robot or for the robot to brake.

\subsection{Visual Prompt Variants}
\label{app:visual_prompt}

Every variant keeps the same eight tools, camera frames, route keyframes, pose readings, node IDs, annotations, observed geometry, and tool behavior. The grounded reference draws labels onto their images in a $5\!\times\!7$-pixel font and shows the map with the map frame's north at the top, a heading arrow, and turn cues relative to the agent's heading; north here refers to the map frame, not to geographic north. Compact glyphs use a $3\!\times\!5$-pixel digit font at the same height and contrast, the heading-up map keeps the same crop and scale, and absolute bearings provide bearing and pose information sufficient to derive the same directions. Labels stay outside the scene content so that no variant changes occlusion.

\section{Evaluation Coverage and Validation}
\label{app:evaluation_validation}

\subsection{R2R Ablation Subset}

The 100-episode ablation set follows the public OpenNav subset~\citep{qiao2025opennav}. All 100 episodes in our runs match the published subset in episode identifier, instruction, and scene, and each matches the R2R-CE v1.3 Val-Unseen release in instruction text, scene, start position, reference path, and geodesic distance. \Cref{tab:subset_audit} compares these episode attributes with the full 1,839-episode split. Numeric entries give mean $\pm$ standard deviation. The absolute standardized mean difference (SMD) compares the subset with its disjoint 1,739-episode complement, using the square root of their equally weighted variances; $D$ is the maximum difference between the subset and full empirical cumulative distributions. Directional mentions count \texttt{left}, \texttt{right}, and forms of \texttt{turn}; reference-path turns are horizontal direction changes of at least $30^\circ$, rather than executed primitive turns. These are interpretable task-complexity proxies, not a complete semantic-complexity measure.

\begin{table}[htbp]
\centering
\small
\caption{R2R ablation-subset coverage and distributions.}
\label{tab:subset_audit}
\setlength{\tabcolsep}{5pt}
\begin{tabular}{lrrrr}
\toprule
Statistic & Full & Subset & $|\mathrm{SMD}|$ & $D$ \\
\midrule
Episodes & 1,839 & 100 & -- & -- \\
Scenes & 11 & 10 & -- & -- \\
Unique reference trajectories & 613 & 97 & -- & -- \\
Geodesic distance (m) & $8.90\pm2.67$ & $8.99\pm3.06$ & 0.033 & 0.072 \\
Instruction length (words) & $26.65\pm11.48$ & $24.68\pm8.28$ & 0.207 & 0.103 \\
Directional mentions & $2.14\pm1.80$ & $2.20\pm1.75$ & 0.034 & 0.040 \\
Reference-path turns & $1.56\pm1.08$ & $1.46\pm1.21$ & 0.095 & 0.100 \\
\bottomrule
\end{tabular}
\end{table}

Path distance and directional mentions have small mean differences; instructions in the subset are shorter and have a narrower length distribution. The subset covers 10 of 11 scenes, omitting a scene containing 18 full-split episodes, and its scene-frequency total-variation distance from the full split is 0.097. In particular, one scene accounts for 3\% of the subset versus 9.63\% of the full split. The Pearson $\chi^2$ statistic of the subset's scene frequencies is 8.14; among 10,000 random 100-episode subsets drawn without replacement, 56.7\% have a statistic at least this large. The observed imbalance is therefore typical of random subsets of this size, but this does not establish that the subset is representative, especially since it was curated. The public OpenNav and v1.3 files also differ in the initial headings of all 100 selected episodes, and we fix them in our evaluation. Thus this comparison establishes coverage and similarity of the listed attributes, not full execution-protocol or performance equivalence. Full-split results remain the basis for benchmark comparisons.

\subsection{Matched ObjectNav Evaluation}

The validation in \Cref{tab:objectnav_matched} runs Qwen-RobotNav 8B and \navgpt on exactly the same 500 episodes for each dataset, with identical scene versions, sensing, motion budgets, collision handling, and stop/success rules. It is a system comparison; the complete system additionally uses its hosted Planner. Selection is fixed before observing either system's outcomes. SR and SPL are percentages. $\Delta$SR vs.\ full is the 500-episode SR minus the full-split SR published for Qwen-RobotNav 8B, in percentage points ($-1.0$ on MP3D, $+1.2$ on HM3D v2); it indicates how closely the selected episodes track the full split. \navgpt has no full-split run, so its entry is a dash.

\begin{table}[htbp]
\centering
\small
\caption{ObjectNav comparison on identical episodes.}
\label{tab:objectnav_matched}
\setlength{\tabcolsep}{5pt}
\begin{tabular}{llrrrr}
\toprule
Dataset & System & $N$ & SR & SPL & $\Delta$SR vs.\ full \\
\midrule
MP3D & Qwen-RobotNav 8B & 500 & 47.8 & 17.5 & $-1.0$ \\
MP3D & \navgpt & 500 & 59.2 & 25.2 & -- \\
HM3D v2 & Qwen-RobotNav 8B & 500 & 72.4 & 33.1 & $+1.2$ \\
HM3D v2 & \navgpt & 500 & 75.3 & 40.8 & -- \\
\bottomrule
\end{tabular}
\end{table}

\subsection{EXPRESS Judge Validation}

We compare the GPT-6 Astra judge with the GPT-4o-mini judge used by prior EXPRESS-Bench results on the same saved answers and final observations, scored against human judgments. \Cref{tab:judge_validation} reports agreement on 50 human-labeled answers: Astra agrees with the human label on 46 and GPT-4o-mini on 36, a 20.0-point difference. Without item-level pairing, the exact McNemar $p$ depends on how the two judges' errors overlap; for these totals it lies between 0.002 and 0.031. This issue applies to open-answer EXPRESS scoring; HM-EQA and MT-HM3D multiple-choice accuracy uses the reference option.

\begin{table}[htbp]
\centering
\small
\caption{EXPRESS judge agreement with human labels.}
\label{tab:judge_validation}
\setlength{\tabcolsep}{5pt}
\begin{tabular}{lrr}
\toprule
Judge & Labeled $N$ & Agreement (\%) \\
\midrule
GPT-4o-mini & 50 & 72.0 \\
GPT-6 Astra & 50 & 92.0 \\
\midrule
Astra $-$ 4o-mini & & +20.0 \\
\bottomrule
\end{tabular}
\end{table}

\end{document}